\documentclass[journal]{IEEEtran}
\usepackage{graphicx}
\usepackage{caption}
\usepackage{amsmath}
\usepackage{bbm}
\usepackage{multirow}
\usepackage[numbers,sort&compress]{natbib}
\usepackage{subfigure}
\usepackage[hidelinks]{hyperref}
\usepackage[linesnumbered,ruled]{algorithm2e}
\usepackage{harpoon}
\usepackage{color}
\ifCLASSINFOpdf
\else
\fi
\begin{document}
%
\title{Goal-oriented Semantic Communications for Robotic Waypoint Transmission: The Value and Age of Information Approach}
%
%
%

\author{Wenchao~Wu, \textit{Student Member, IEEE}
        Yuanqing~Yang,
        Yansha~Deng, \textit{Senior Member, IEEE}
        and~A.~Hamid~Aghvami, \textit{Life Fellow, IEEE}
\thanks{Manuscript received 16 December 2023;revised 09 May 2024; accepted 26 June 2024. This work was supported by the Engineering and Physical Sciences Research Council (EPSRC), U.K., under Grant EP/W004348/1.This work is also a contribution by Project REASON, a UK government-funded project under the FONRC sponsored by the DSIT. An early version of this paper was presented in part at the 2024 IEEE International Conference on Communications in June 2024. The associate editor coordinating the review of this article and approving it for publication was X. Cao. (Corresponding author: Yansha Deng)

The authors are with the Department of Engineering, King’s College London, Strand, London WC2R 2LS, U.K. (e-mail: wenchao.wu@kcl.ac.uk; yuanqing.yang@kcl.ac.uk; yansha.deng@kcl.ac.uk; hamid.aghvami@kcl.ac.uk)
.}
}

\maketitle

\begin{abstract}
The ultra-reliable and low-latency communication (URLLC) service of the fifth-generation (5G) mobile communication network struggles to support safe robot operation. Nowadays, the sixth-generation (6G) mobile communication network is proposed to provide hyper-reliable and low-latency communication to enable safer control for robots. However, current 5G/ 6G research mainly focused on improving communication performance, while the robotics community mostly assumed communication to be ideal. To jointly consider communication and robotic control with a focus on the specific robotic task, we propose goal-oriented semantic communication in robotic control (GSRC) to exploit the context of data and its importance in achieving the task at both transmitter and receiver. At the transmitter, we propose a deep reinforcement learning algorithm to generate optimal control and command (C\&C) data and a proactive repetition scheme (DeepPro) to increase the successful transmission probability. At the receiver, we design the value of information (VoI) and age of information (AoI) based queue ordering mechanism (VA-QOM) to rank the queue based on the semantic information extracted from AoI and VoI. The simulation results validate that our proposed GSRC framework achieves a 91.5$\%$ improvement in the mean square error compared to the traditional unmanned aerial vehicle control framework.

\end{abstract}


%
\IEEEpeerreviewmaketitle

\section{Introduction}
%
%
%
%
\IEEEPARstart{T}{he} field of robotics has experienced significant advancements in the past few decades, leading to widespread applications, such as agriculture \cite{Robotics1}, education \cite{Robotics2}, and surgery \cite{Robotics3}. Wireless connectivity is important for the control of mobile robots (i.e., ground vehicles and unmanned aerial vehicles (UAVs)) \cite{Wireless_communication1}, especially for tasks that require remote control and sensing \cite{Wireless_communication2}. Remote control of robots can benefit from the development of the fifth-generation (5G) mobile communication network \cite{5G1}, due to its ultra-reliable low latency communication (URLLC) service with a latency requirement as low as 1 ms \cite{URLLC1, URLLC2, URLLC3,5G2}. Unfortunately, this URLLC service defined in 5G is still unable to support the real-time safe operation of mobile robots \cite{Drawback1}. In the context of high-speed mobile robots, even a mere 1 ms transmission delay can result in significant movement that cannot be neglected. In such case, the mobile robot may potentially collide with obstacles or pose a risk to the surrounding objects \cite{Drawback2}.

\par Compared to 5G, the sixth-generation (6G) mobile communication network \cite{6G1} aims to provide a more stringent service requirement, known as hyper-reliable and low-latency communication, which demands a peak data rate ranging from 50 Gbps to 200 Gbps, latency within 0.1ms to 1ms, and reliability ranging from 1-$10^{-5}$ to 1-$10^{-7}$ \cite{6G2}. 6G technology is expected to facilitate the remote control of mobile robots with accuracy, robustness, energy efficiency, and safety \cite{Drawback1}. Until now, existing research has been mainly focused on improving the communication performance, such as the reliability \cite{6G_communication2}, the data rate \cite{6G_communication1}, and the latency \cite{6G_communication3}.

\par On one hand, existing wireless communication works tend to ignore the actuation performance of robotics, and only focus on improving the communication performance. On the other hand, existing robotics papers \cite{Rob_Control1, Rob_Control2, Transmitter} mostly assumed wireless communication to be ideal without packet loss or latency. Even in the case where robots are controlled remotely, as demonstrated in \cite{Transmitter}, the transmitter (remote controller) is responsible for periodically publishing control commands, and the receiver (mobile robot) stores the commands in a queue based on their arrival time. Therefore, there is a big research gap on how to jointly consider and optimize wireless communication and robotic control with a focus on specific robotic tasks, such as tracking, navigation, etc.

\par To fill this gap, goal-oriented communication has been proposed in \cite{TOSA1} to design the communication system/ network with a focus on the effectiveness of communication in accomplishing a specific goal. In the robotics field, the effectiveness of controlling the robots wirelessly can be measured by effectiveness level performance metrics, such as the mean square error (MSE) \cite{task_per1} and the tracking accuracy \cite{task_per2}. In \cite{AI_TO3}, the authors proposed a deep reinforcement learning (DRL) method to choose the optimal segment length and sampling rate for the physical world trajectory transmission of a robotic arm, with the aim to improve the tracking error measured by the MSE between the trajectory of the device and its digital model. Even though \cite{AI_TO3} optimized goal-oriented performance (i.e., tracking error), the meaning and importance/ significance behind the meaning of bits are completely ignored.

\par To enable more accurate, context-aware, and efficient communication, semantic communication is introduced to solve the semantic-level problem \cite{SC}. Different from the well-studied technical-level problem focusing on how to reliably transmit bits, the semantic-level problem considers how to convey semantic information precisely and serves as the foundation for the effectiveness-level problem \cite{TO1}. Semantic information has two main definitions. One is defined as the meaning of the information, which pertains to what the information is intended to express and how it relates to the world around it, entailing a focus on syntax \cite{TO1}. This definition has mainly considered for text data type, where the bilingual evaluation understudy (BLEU) score is introduced to evaluate the sentence similarity for machine translation \cite{SA1}, signal-to-distortion ratio (SDR) and perceptual evaluation of speech distortion (PESQ) are used to evaluate speech signal reconstruction \cite{SA2}.

\par The second definition pertains to the significance and importance of information \cite{SA4}, which evaluates the importance of extracted semantic information in accomplishing a specific task and is closely coupled with the considered task. Depending on the task/ goal, the significance of information can be evaluated based on various new metrics, including the age of information (AoI), the value of information (VoI), just noticeable difference \cite{TOSA1}, the skeleton of the avatar \cite{TOSA3}, and et. The AoI is defined as the duration elapsed since the information was first generated and is widely used to effectively characterize latency in the real-time status update system \cite{AoI2}. In \cite{AoI1}, the authors proposed a semantics-aware approach to address the problem of active fault detection in an IoT scenario. A stochastic approximation algorithm was employed to jointly consider the importance and purpose of the information to detect faults and how to acquire fresh information more effectively. In this approach, the AoI was utilized to evaluate the freshness of the sensor information, and the semantic information was measured by a function as VoI, which was related to the AoI and the entropy of the probability distribution over the operational status. The concept of VoI is commonly defined as the value/ significance of the information in completing communication tasks and can be utilized as a metric to quantify the significance of semantic information \cite{VoI1,VoI2}. Depending on the task/ goal, VoI has been represented based on various metrics, such as the importance of the extracted features to the accurate classification \cite{VoI1}, the cost for wrong actuation \cite{VoI3}, and structural index similarity (SSIM) \cite{SA3}.

\par To incorporate semantic level consideration into the task-oriented communication framework, a generic task-oriented and semantics-aware (TOSA) communication framework was proposed in \cite{TOSA1}. This framework jointly considered the semantic level information and effectiveness level performance metrics, for different tasks with various data types. Although TOSA has been applied in various domains \cite{TOSA2,TOSA3,TOSA4}, there is little research on the command and control data for robotics. In \cite{TOSA5}, the TOSA communication framework was studied for the downlink control and command (C\&C) signal transmission, which characterized semantic information as a function of AoI and the similarity between adjacent C\&C signals. This study effectively reduced redundant packet transmissions by utilizing a DRL-based TOSA optimization method. However, it lacks the design for effectiveness level metrics and the receiver design. 

\par To fill this gap, we propose goal-oriented and semantic communications in robotic control (GSRC) in a real-time UAV waypoint transmission task. Compared to traditional transmitter/receiver design and previous works \cite{Transmitter,AI_TO3,AoI1,VoI3,TOSA5}, we design the DRL at the transmitter to generate the optimal C\&C data to accomplish the task, adopt the proactive repetition scheme for C\&C data transmission, order the queue of received C\&C data based on their semantic information extracted from the AoI and the VoI, and utilize the MSE as our effectiveness level metric. To the best of our knowledge, our contributions can be summarized as follows:
\begin{itemize}
    \item[1)] We consider a scenario involving a BS and a UAV with a dynamic communication environment. In this setup, the BS acts as the transmitter, periodically transmitting C\&C data to the UAV to manoeuvre it to fly along a trajectory. Our goal is to collaboratively consider the communication and control procedures on both the BS and UAV sides, with the aim of minimizing the positional error between the UAV's actual flight trajectory and target trajectory.
    \item[2)] We propose a GSRC framework, which includes DRL-based C\&C data generation with the proactive scheme for repetition (DeepPro) and the VoI and the AoI based queue ordering mechanism (VA-QOM). At the semantic level, the GSRC employs the DeepPro on the BS side, facilitating the generation of more appropriate C\&C data to accomplish the task and elevating the probability of successful transmission compared to conventional frameworks. Meanwhile, on the UAV side, the VA-QOM is deployed by utilizing the combined function of AoI and VoI to extract the semantic information of received C\&C data in the queue, which can measure the importance of each C\&C data at the present moment and has the potential to be extended to other time-sensitive control applications. At the effectiveness level, we choose the MSE between the actual flight trajectory of the UAV and the target one as the performance metric.
    \item[3)] To evaluate the effectiveness of the proposed GSRC framework, we implemented it in a real-time UAV waypoint transmission task and compared it with the traditional UAV control framework via simulation. The simulation results indicate that the proposed GSRC framework achieves 91.5$\%$ improvement in the mean square error compared to the traditional UAV control framework.
\end{itemize}

\par The rest of the paper is organized as follows: In Section $\mathrm{\uppercase\expandafter{\romannumeral2}}$, we present the system model and problem formulation. Section $\mathrm{\uppercase\expandafter{\romannumeral3}}$ and Section $\mathrm{\uppercase\expandafter{\romannumeral4}}$ introduce the design of the traditional UAV control framework and the proposed GSRC framework, respectively. In Section $\mathrm{\uppercase\expandafter{\romannumeral5}}$, the simulation results is presented. Finally, Section $\mathrm{\uppercase\expandafter{\romannumeral6}}$ concludes this paper.

\section{System Model and Problem formulation}
In this section, we introduce a real-time UAV waypoint transmission task, where a BS controls a UAV to fly alongside a trajectory in real time by sending C\&C data. Subsequently, we model the communication environment for downlink C\&C data transmission. Based on that, we present the problem formulation and the evaluation metric.

\subsection{Real-time UAV Waypoint Transmission Task}

\begin{figure}[htp]
    \centering
    \includegraphics[width=0.9\linewidth]{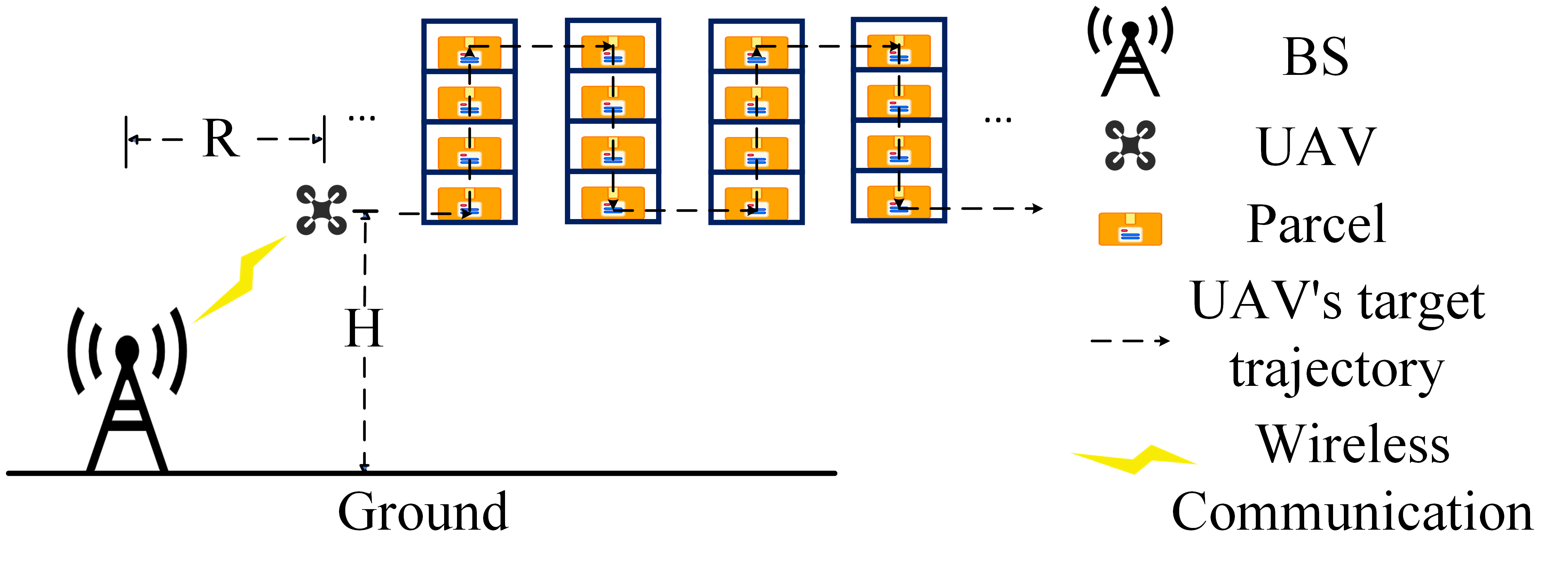}
    \caption{User case.}
    \label{User case}
\end{figure}
We consider a real-time UAV waypoint transmission task, where a BS serves as a remote controller to generate C\&C data to control a UAV to fly alongside a trajectory in real time. The trajectory consists of a series of waypoints, which are specific geographic locations and serve as reference points for navigation. In the task, the BS only has knowledge of the UAV's target waypoint in the next time slot. As a result, the BS must transmit the corresponding C\&C data in real time. By executing the received C\&C data, the UAV flies across various waypoints at different timestamps. One user case is shown in Fig. \ref{User case}, where the BS needs to control the UAV in real time to scan QR codes on the parcels located at various positions. To effectively complete the task within the time constraint, the UAV aims to reach each waypoint on time.

\subsection{Channel Model}
We assume that the BS remains stationary throughout the communication, and the UAV flies within a circular horizontal disk with a radius of R and a height of H, as shown in Fig. \ref{User case}.
\par Considering both line-of-sight (LoS) and non-line-of-sight (NLoS) conditions for flying UAVs, we utilize free-space path loss and Rayleigh fading to model the path loss from the BS to the UAV as
\begin{equation}
     h =
    \left\{
    \begin{aligned}
        &{\left( \frac{4 \pi d f_c}{c} \right)}^{\alpha} \eta_{\mathrm{LoS}}\beta, \quad P_{\mathrm{LoS}}\\
        &{\left( \frac{4 \pi d f_c}{c} \right)}^{\alpha} \eta_{\mathrm{NLoS}}\beta, \quad P_{\mathrm{NLoS}}=1-P_{\mathrm{LoS}},
    \end{aligned} 
    \right
    .
\end{equation}
where $d$ is the distance between the UAV and the BS, $f_c$ is the transmission frequency, $c$ is the speed of light, $\alpha$ is the path loss exponent, $\beta$ is the Rayleigh small-scale fading which follows $\mathcal{CN}\left(0,1\right)$, and $\eta_{\mathrm{LoS}}$ and $\eta_{\mathrm{NLoS}}$ is the path loss coefficients of LoS and NLoS cases, respectively. In Eq. (1), the LoS probability $P_{\mathrm{LoS}}$ is derived as
\begin{equation}
    P_{\mathrm{LoS}}=\frac{1}{1+ae^{-b\left( \theta-a \right)}},
\end{equation}
where $\theta=\frac{180}{\pi}\arcsin\frac{\mathrm{H}}{d}$ is the evaluation angle of the UAV, $\mathrm{H}$ is the flight height of the UAV, and $a$ and $b$ are positive constants related to the environment. Based on Eq. (1), the signal-to-noise ratio (SNR) can be formulated as
\begin{equation}
    \mathrm{SNR}= \frac{\mathrm{P}h}{\sigma^2},
\end{equation}
where $\mathrm{P}$ is the transmit power, $h$ is the channel gain, and $\sigma^2$ is the Additive White Gaussian Noise (AWGN) power. Then, we can derive the transmission time for each C\&C data as
\begin{equation}
    t^\mathrm{CC}=\frac{\mathrm{N_{CC}}}{B\log_{2}\left( \mathrm{SNR+1} \right)},
\end{equation}

where $\mathrm{N_{CC}}$ is the size of each C\&C data, and $B$ is bandwidth. In downlink transmission, the UAV can only successfully decode the C\&C data when the SNR of C\&C transmission exceeds a specific threshold denoted as $\gamma_\mathrm{th}$. The parameter that determines whether C\&C data can be successfully decoded is expressed as
\begin{equation}
    \delta^\mathrm{CC}=\begin{cases}
        0,\quad \mathrm{SNR} \leq \gamma_\mathrm{th}\\
        1,\quad \mathrm{SNR}>\gamma_\mathrm{th}.
    \end{cases}
\end{equation}
The C\&C data can be successfully decoded if $\delta^\mathrm{CC}=1$, otherwise, the C\&C data can not be decoded. Meanwhile, if one C\&C data is successfully decoded, we assume that it is not distorted and can be accurately recovered.

\subsection{The General Communication and Control Process}
To provide an overview of the general communication and control process for the task, we introduce the C\&C data transmission protocol and the queue employed to store the received C\&C data. 

\subsubsection{C\&C Data Transmission}
The BS starts to generate and transmit the C\&C data periodically at the beginning of $i^\mathrm{th}$ transmission time interval (TTI) $t_{i-1}$, where $i \in \left\{1,2,..., N^\mathrm{TTI}\right\}$ and $N^\mathrm{TTI} \in \mathbbm{N}$ is the maximum value of $i$. The time duration of each TTI is fixed as a constant $T$. If the UAV successfully receives and decodes the C\&C data $\boldsymbol{m}_i$, it will send an acknowledgement (ACK) to the BS. Otherwise, it will send a negative acknowledgement (NACK) to the BS. The parameter determining to send ACK/ NACK is defined as
\begin{equation}
    \delta^\mathrm{ACK}_i=\begin{cases}
        0,\ \delta^\mathrm{CC}_i=0\\
        1,\ \delta^\mathrm{CC}_i=1,
    \end{cases}
\end{equation}
where $\delta^\mathrm{CC}_i$ is the transmission result (success or failure) for $\boldsymbol{m}_i$. If $\delta^\mathrm{ACK}_i=1$, the UAV sends ACK. Otherwise, the UAV sends NACK.
\subsubsection{Queue}
The queue of a UAV is a crucial component of the UAV's control system, serving as a structured list of commands and instructions to be executed. When the UAV successfully receives and decodes one C\&C data, it will store the C\&C data in the queue with a predefined size $Q_\mathrm{max}$. The size $Q_\mathrm{max}$ is not fixed in all scenarios and can be customized for the needs of a particular task. If the number of received C\&C data exceeds $Q_\mathrm{max}$, the C\&C data with lower priority will be discarded. The priority can be measured by various metrics, such as the arrival time and the semantic information. The UAV ranks the C\&C data in the queue based on their priorities and executes the C\&C data according to the order. This adaptability allows for efficient task management and execution, ensuring that the UAV can effectively handle different scenarios and workloads as needed. 

\subsubsection{The Overall Process}
We assume that the time duration of the whole process is $\left[0,TN^\mathrm{TTI} \right]$. The UAV's actual position and target position at $t\in \left[0,TN^\mathrm{TTI}\right]$ can be represented as $\boldsymbol{p}_{\frac{t}{T}}$ and $\boldsymbol{g}_{\frac{t}{T}}$, respectively. At the beginning of $i^\mathrm{th}$ TTI $t_{i-1}=\left(i-1\right)T$, the BS generates and transmits the C\&C data $\boldsymbol{m}_i$ with the execution time of $T$, expressed as
\begin{equation}
    \boldsymbol{m}_i=\left( v_i^x, v_i^y, v_i^z \right),
\end{equation}
where $v_i^x$, $v_i^y$, and $v_i^z$ are the UAV's planned velocity on the x-axis, y-axis, and z-axis for the $i^\mathrm{th}$ TTI, respectively. Upon successfully receiving and decoding the C\&C data $\boldsymbol{m}_i$, the UAV sends an ACK back ($\delta^\mathrm{ACK}_i=1$) to the BS. Otherwise, the UAV sends a NACK ($\delta^\mathrm{ACK}_i=0$) instead. When the UAV successfully receives and decodes $\boldsymbol{m}_i$, it will store it in the queue and rank all the received C\&C data in the queue. By executing the C\&C data in the queue according to their order, the UAV's position is updated. It is important to note that, at the end of $i^\mathrm{th}$ TTI $t_{i}=iT$, the UAV sends its real-time position $\boldsymbol{p}_{i}=\left(x_i,y_i,z_i\right)$ back to the BS, where $x_{i}$, $y_{i}$, and $z_{i}$ are the UAV's position on the x-axis, y-axis, and z-axis, respectively. To focus on the downlink transmission, we assume the uplink transmission (i.e., ACK, NACK, and $\boldsymbol{p}_{i}$) from the UAV to the BS is ideal without packet loss or delay.

\subsection{Problem Formulation and Performance Metric}
Due to the delay and packet loss in the downlink transmission of C\&C data, there is often a discrepancy between the UAV's actual trajectory and the UAV's target trajectory. Consequently, we utilize the mean square error (MSE) to measure the gap between the two trajectories. Specifically, we aim to minimize the MSE between the positions of the actual trajectory and their corresponding positions of the target trajectory at the end of each TTI. We assume that each TTI can be uniformly divided into $N^\mathrm{M}$ time duration. For $j^\mathrm{th}$ time duration in $i^\mathrm{th}$ TTI where $j\in \left\{1,...,N^\mathrm{M}\right\}$, its timestamp is represented as $\left[\left(i-1\right)T+\frac{T}{N^\mathrm{M}}j\right]$. In this way, we can obtain the corresponding positions of the UAV's actual trajectory $\boldsymbol{p}_{\left(i-1+\frac{j}{N^\mathrm{M}}\right)}$ and the UAV's target trajectory $\boldsymbol{g}_{\left(i-1+\frac{j}{N^\mathrm{M}}\right)}$. When $N^\mathrm{M}$ is infinity, the average MSE of these two positions represents the MSE of the two trajectories. As a result, the problem can be formulated as 
\begin{equation}
\begin{aligned}
    &\mathcal{P}1:\min \lim_{N^\mathrm{M}\rightarrow \infty}\frac{1}{N^\mathrm{TTI}N^\mathrm{M}}\sum_{\substack{i=1}}^{\substack{N^\mathrm{TTI}}} \sum_{\substack{j=1}}^{\substack{N^\mathrm{M}}}
     ||&\boldsymbol{p}_{\left(i-1+\frac{j}{N^\mathrm{M}}\right)}-\\
     &&\boldsymbol{g}_{\left(i-1+\frac{j}{N^\mathrm{M}}\right)}||^2
     \\
     &\mathrm{s.t.}\ N^\mathrm{M} \in \mathbbm{N},
\end{aligned}
\end{equation}
where $||.||$ is the Frobenius norm.
\par Similar to the problem formulation, we utilize the MSE between the UAV's actual trajectory and target trajectory as the performance metric. To simplify the calculation, instead of obtaining the positions of infinity time duration, we set $N^\mathrm{M}$ as a fixed constant and the performance metric can be calculated as
\begin{equation}
    \mathrm{MSE}=\frac{1}{N^\mathrm{M}N^\mathrm{TTI}}\sum_{i=1}^{N^\mathrm{TTI}}\sum_{j=1}^{N^\mathrm{M}}||\boldsymbol{p}_{\left( i-1+\frac{j}{N^\mathrm{M}} \right)}-\boldsymbol{g}_{\left( i-1+\frac{j}{N^\mathrm{M}} \right)}||^2.
\end{equation}
\section{Traditional UAV Control Framework}
\begin{figure}[!t]
    \centering
        \includegraphics[width=0.9\linewidth]{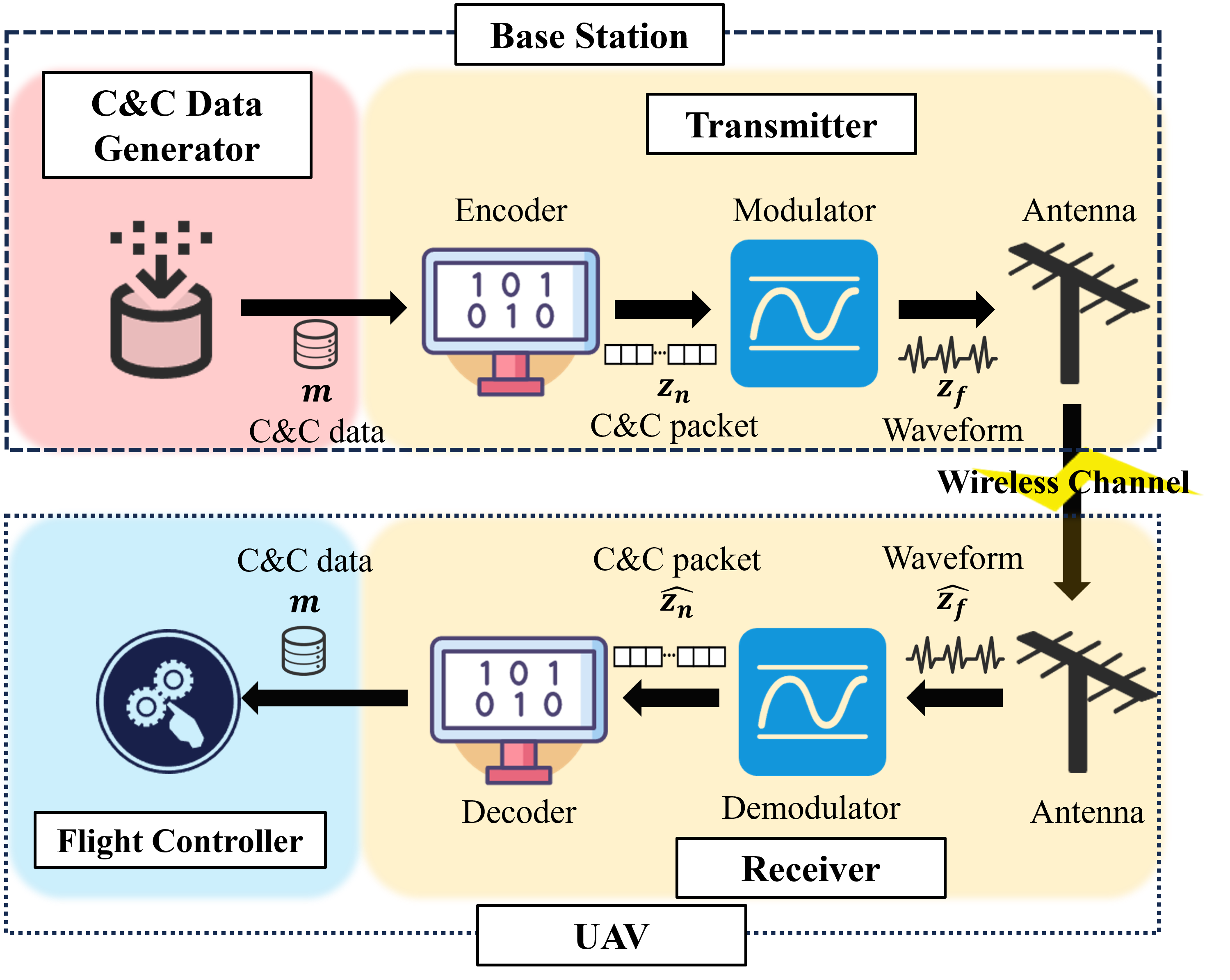}
        \caption{Traditional UAV control framework}
        \label{TUCF}
\end{figure}

As our work is the first work that considers a practical system model with C\&C generation, C\&C data wireless transmission, and C\&C queue at the receiver, taking into account the content of C\&C data, rather than simply treating them as bits, there is no other state-of-the-art research that can be compared with our solutions in this practical system model. Due to this reason, we introduce the traditional UAV control framework as the baseline scheme in this section to reflect how the current practical wireless UAV system works, such as in \cite{Transmitter}. As shown in Fig. \ref{TUCF}, it includes the design at the BS side and the UAV side. The BS consists of two main components: the C\&C data generator and the transmitter. The C\&C data generator is responsible for generating the C\&C data at the beginning of each TTI. The transmitter includes an encoder, a modulator, and an antenna, which is used to convert the C\&C data to the waveform and broadcast it to the wireless channel.
\begin{algorithm}[!t]
\DontPrintSemicolon
    \SetAlgoLined
    \caption{Traditional UAV Control Framework}
    \KwIn{$a$, $b$, $f_c$, $\alpha$, $\eta_{\mathrm{LoS}}$, $\eta_{\mathrm{NLoS}}$, $\sigma^2$, $\mathrm{N_{CC}}$, $B$, and $\gamma_\mathrm{th}$.}
    \KwOut{UAV actual position set $\mathcal{P}$ with its corresponding time set $\mathcal{T}$.}
    Initialization: The UAV's initial position $\boldsymbol{p}_0$ with its corresponding time $t_0$, UAV's received C\&C data set $\mathcal{C}$.\\
    \For{$i\leftarrow 1$ to $N^\mathrm{TTI}$}
    {
        Update the UAV's position $\boldsymbol{p}_{i-1}$.\\
        Initialize UAV's received C\&C data set in $i^\mathrm{th}$ TTI $\mathcal{C}_i$ to $\emptyset$.\\
        Generate $\boldsymbol{m}_i$.\\
        Calculate $P_{\mathrm{LoS}}$ using Eq. (2).\\
        Calculate $h_i$ using Eq. (1).\\
        Calculate SNR using Eq. (3).\\
        Calculate $t_{i}^\mathrm{CC}$ using Eq. (4).\\ 
        Calculate $\delta^\mathrm{CC}_i$ using Eq. (5).\\
        \If{$\delta^\mathrm{CC}_i=1$}
            {
                Add $\boldsymbol{m}_i$ in $\mathcal{C}$.\\
            }
        \For{all C\&C data in $\mathcal{C}$}
            {
                \If{the C\&C data arrived in $\left[t_{i-1},t_{i}\right)$}
                    {
                        Add this C\&C data in $\mathcal{C}_i$.\\
                    }
            }
        Reorder the C\&C data in $\mathcal{C}_i$ by their arrival time.\\
        \For{all C\&C data in $\mathcal{C}_i$}
        {
            Update the UAV's position.\\
        }
    }
\end{algorithm}
\par At the UAV side, it comprises a receiver and a flight controller. The receiver consists of an antenna, a demodulator, and a decoder, which is able to recover the C\&C data from the received waveform if the C\&C data can be successfully decoded (i.e., $\mathrm{SNR}>\gamma_\mathrm{th}$). In the traditional UAV control framework, the queue size $Q_\mathrm{max}$ is set to one, and the queue is ordered by the arrival time of the C\&C data. Therefore, only the newly arrived C\&C data $\boldsymbol{m}$ is stored in the queue, and delivered to the flight controller. The UAV's flight controller is able to convert the received C\&C data $\boldsymbol{m}$ into direct commands (velocity $\boldsymbol{v}$ and the execution time $\tau$), which are used to control the UAV to accomplish the specific task.
\begin{figure}[!t]
    \centering
        \includegraphics[width=0.9\linewidth]{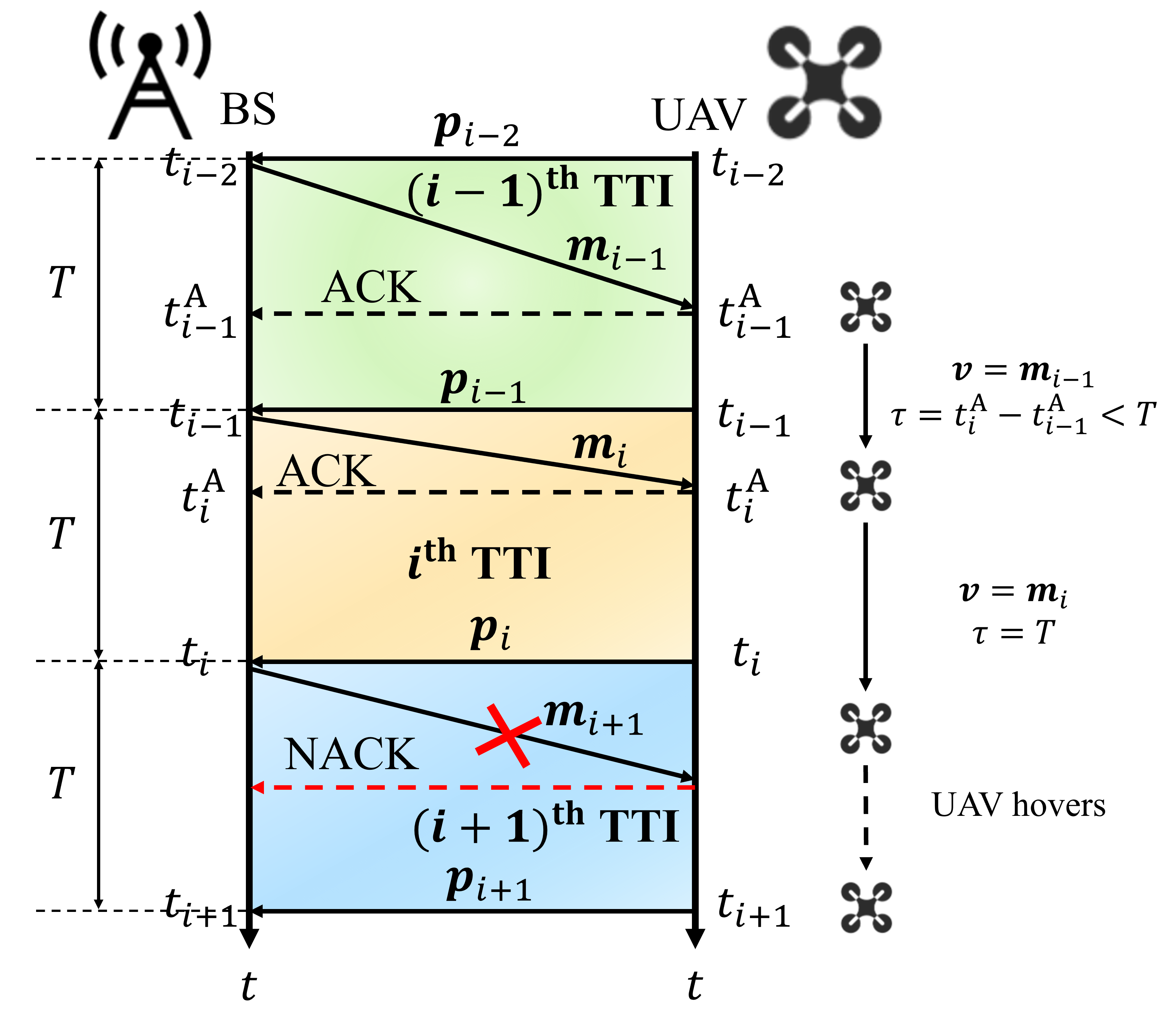}
        \caption{Exemplar timeline in the traditional UAV control framework.}
        \label{TUCF_timeline}
\end{figure}
\par By assuming the C\&C data processing time at both the BS and the UAV sides can be neglected, we plot an exemplar timeline of the BS and the UAV in the traditional UAV control framework shown in Fig. \ref{TUCF_timeline}. At the beginning of $\left(i-1\right)^\mathrm{th}$ TTI $t_{i-2}$, the BS transmits the C\&C data $\boldsymbol{m}_{i-1}$ to the UAV. Once the UAV successfully decodes $\boldsymbol{m}_{i-1}$ at $t_{i-1}^\mathrm{A}$, it will fly in a constant speed $\boldsymbol{v}=\boldsymbol{m}_{i-1}$. If the UAV successfully decodes new C\&C data $\boldsymbol{m}_i$ when executing the ongoing C\&C data $\boldsymbol{m}_{i-1}$ at $t_i^\mathrm{A}$, it will discard the ongoing C\&C data $\boldsymbol{m}_{i-1}$ and start to execute the new C\&C data $\boldsymbol{m}_{i}$. After finishing executing the C\&C data $\boldsymbol{m}_{i}$, the UAV remains hovering until one new C\&C data is successfully decoded.
\par The design of the traditional UAV control framework is shown in $\mathbf{Algorithm\ 1}$. At the beginning of $i^\mathrm{th}$ TTI $t_{i-1}$, the UAV's position $\boldsymbol{p}_{i-1}$ is updated. By utilizing the channel model, we can obtain the transmission result $\delta_i$ and transmission time $t_{i}^\mathrm{CC}$ for $\boldsymbol{m}_i$. Subsequently, all the successfully decoded C\&C data in $i^\mathrm{th}$ TTI $\left[t_{i-1},t_{i}\right)$ can be obtained, which can be utilized to update the UAV's position. The algorithm iterates until all the real-time C\&C data are transmitted.  

\section{The Proposed GSRC Framework}
\begin{figure*}[htbp]
    \centering
    \includegraphics[width=0.65\textwidth]{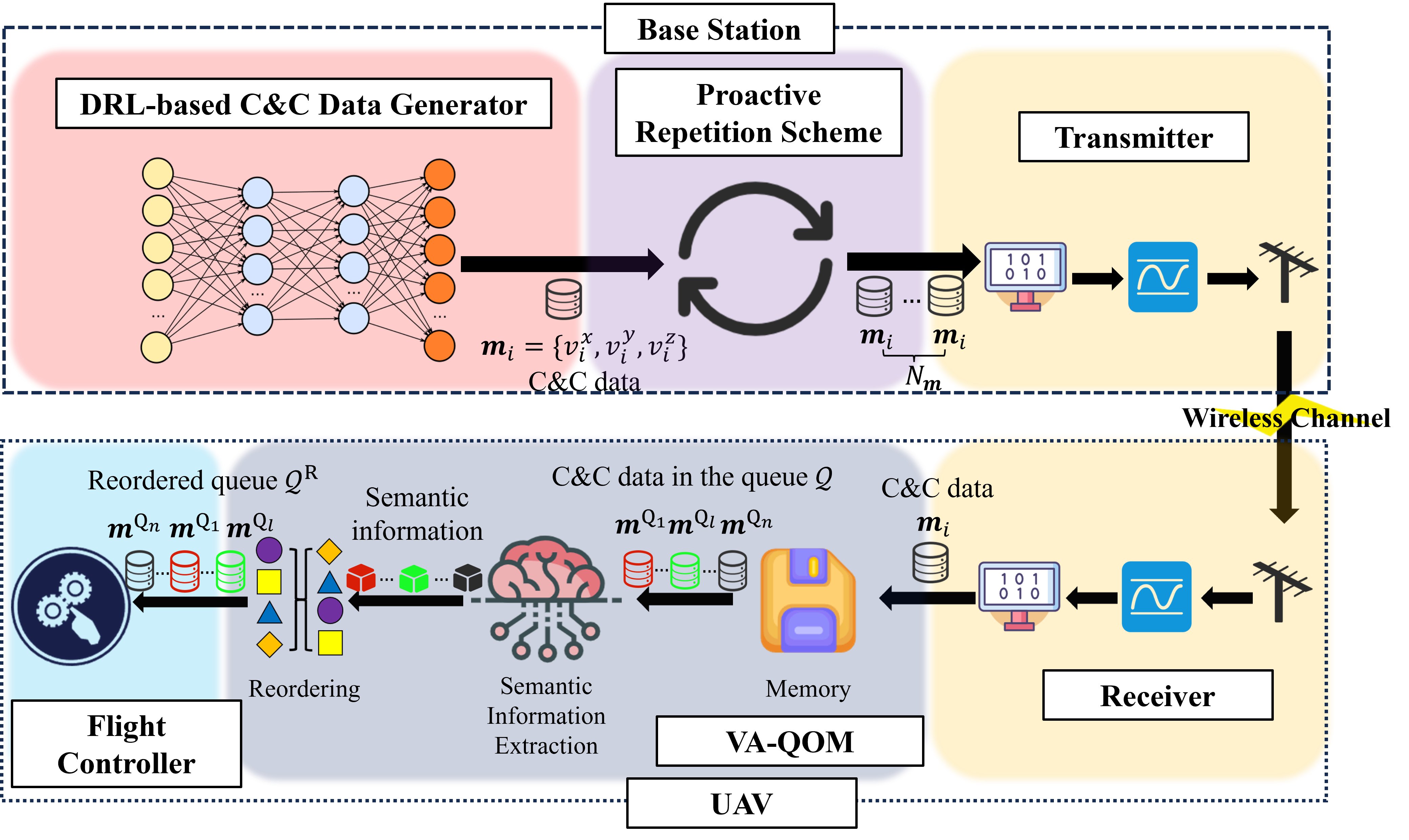}
    \caption{The proposed GSRC framework}
    \label{GSRC}
\end{figure*}
In this section, we propose the GSRC framework which incorporates the semantic level design into the effectiveness level metric. As shown in Fig. \ref{GSRC}, compared with the traditional UAV control framework, our proposed GSRC framework includes DeepPro at the BS, and the VA-QOM at the UAV.

\subsection{DeepPro at the BS}
At the BS, we design a DeepPro scheme, which is expected to generate optimal C\&C data for the real-time UAV waypoint transmission task via the DRL-based C\&C data generator and increase the probability of successful transmission for each C\&C data via the proactive repetition transmission. 

\subsubsection{DRL-based C\&C Data Generator}
To generate optimal C\&C data at the beginning of each TTI, we need to obtain the influence of this C\&C data on the task performance, which is related to the gap between the UAV's actual positions and the UAV's target positions of future TTIs. We define the state of the $i^\mathrm{th}$ TTI as $S_i=\left\{x_{i-1},y_{i-1},z_{i-1},t_{i-1}\right\}$, where $x_{i-1}$, $y_{i-1}$, and $z_{i-1}$ are the UAV's position on the x-axis, y-axis, and z-axis at the beginning of $i^\mathrm{th}$ TTI $t_{i-1}=\left(i-1\right)T$, respectively. We aim to tackle the problem of optimizing the C\&C data generation defined by the action parameter $A_i=\left\{v_i^x, v_i^y, v_i^z\right\}$ at $i^\mathrm{th}$ TTI, where $v_i^x\in V^\mathrm{X}$, $v_i^y\in V^\mathrm{Y}$, and $v_i^z\in V^\mathrm{Z}$ are the UAV's planned velocity in x-axis, y-axis, and z-axis, respectively. At the beginning of $i^\mathrm{th}$ TTI $t_{i-1}$, the agent at the BS makes the decision by accessing all prior historical observations $\left\{U_{1},...,U_{i^\mathrm{'}},...,U_{i-1}\right\}$ for all previous TTIs $i^\mathrm{'} \in \left\{1,...,i-1\right\}$, where the observation 
$U_{i^\mathrm{'}}$ consists of the following variables: the UAV's actual position $\boldsymbol{p}_{i^\mathrm{'}-1}$ and the UAV's target position $\boldsymbol{g}_{i^\mathrm{'}-1}$ at $t_{i^\mathrm{'}-1}$. It is noted that the observation in each TTI $i$ includes all histories of such measurements and
past actions, which is denoted as $O_i=\left\{A_1,U_1,A_2,U_2,...,A_{i-1},U_{i-1}\right\}$.
\par To obtain the impact of the C\&C data $\boldsymbol{m}_i$ generated at the start of $i^\mathrm{th}$ TTI $t_{i-1}$, the BS aims at minimizing the distance between the UAV's actual positions and the UAV's target positions in future TTIs. The optimization relies on the selection of parameters in $A_i$ according to the current observation history $O_i$ with respect to the stochastic policy $\pi$, which can be formulated as
\begin{equation}
     \mathcal{P}2:\quad \max_{\pi\left( A_i|O_i \right)} \sum_{{b=i}}^{{\infty}}\gamma^{b-i}\mathbbm{E}_{\pi}[-||\boldsymbol{p}_{b}-\boldsymbol{g}_{b}||],
\end{equation}
where $\gamma \in \left(0,1\right]$ is the discount factor for the performance accrued in future TTIs, and $||\boldsymbol{p}_{b}-\boldsymbol{g}_{b}||$ is the distance between the UAV's actual position $\boldsymbol{p}_{b}$ and target position $\boldsymbol{g}_{b}$ at the end of $b^\mathrm{th}$ TTI $t_{b}$. Based on that, the reward $R_i$ is formulated as
\begin{equation}
     R_i=-||\boldsymbol{p}_{i}-\boldsymbol{g}_{i}||.
\end{equation}
It is noticed that the smaller the distance is after selecting one action, the higher the reward this action has. 
\par The C\&C data selection process at the BS is the Markov Decision Process because the state $S_i$ is only related to its previous state and past action. We define $\mathcal{S}$ as a set of actual environment’s states with variables of the UAV's positions at a certain time $\left\{x,y,z,t\right\}$ and channel parameters. The fact is that the BS cannot fully observe actual states in $\mathcal{S}$, while it can only access variables of $\left\{x,y,z,t\right\}$ without channel parameters. Meanwhile, the packet loss or delay decided by the channel parameters influences the transition between any two successive states. Since the channel parameter cannot be assessed by the BS, the process of state transition is not fully observed, which introduces a generally intractable Partially Observable Markov Decision Process (POMDP) problem.
\par To tackle this problem, we design the deep-Q network (DQN) algorithm, as one powerful DRL approach. The agent of the DQN algorithm is deployed at the BS to choose optimal actions by exploring the environment. During the training process, learning takes place over multiple episodes, with each episode including $N^\mathrm{TTI}$ TTIs.
\par From the environment perspective, when it receives an action $A_i$ from the agent which is in state $S_i$, it responds by providing the action's reward $R_i$ and the agent's next state $S_{i+1}$. From the agent perspective, it receives the sample $\left(S_i,A_i, R_i, S_{i+1}\right)$ and stores it in the replay memory. The replay memory serves as a buffer that stores past experiences for the agent to learn from. Once the replay memory stored enough samples, the agent randomly selects a batch of samples for training. During the training process, the current state $S_i$ and its corresponding action $A_i$ are inputted to the Q-network, which calculates the predicted value using the function $Q\left(S_i,A_i;\boldsymbol{\theta}_i\right)$. Here, $\boldsymbol{\theta}_i$ represents the parameter vector of the Q-network. Simultaneously, the next state $S_{i+1}$ and the reward $R_i$ are passed to the target Q-network, which is a separate network used for calculating the target values. Then, they pass the results to the loss function calculation parts, and the gradient of the loss function is calculated and utilized to update the parameter vector in the Q network. The gradient of the loss function is given as
\begin{multline}
    \nabla L\left( \boldsymbol{\theta}_i \right)=\mathbbm{E}_{S_i,A_i,R_i,S_{i+1}}
    \left[ R_i+\gamma \max_{A} Q\left( S_{i+1},A;\boldsymbol{\theta}^-_i \right) \right.
    \\
    \left. -Q\left( S_{i},A_i;\boldsymbol{\theta}_i \right) \nabla_{\boldsymbol{\theta}} 
     Q\left( S_i,A_i;\boldsymbol{\theta}_i \right) \right],
\end{multline}
where $\boldsymbol{\theta}_i$ is updated by
\begin{equation}
\boldsymbol{\theta}_{i+1}=\boldsymbol{\theta}_i-\lambda_{\mathrm{RMS}} \nabla L\left( \boldsymbol{\theta}_i \right),
\end{equation}
where $\lambda_{\mathrm{RMS}}$ is the learning rate of RMSprop. In the training process, it is noted that the parameter vector $\boldsymbol{\theta}^-$ in the target Q-network is updated by copying the parameter vector $\boldsymbol{\theta}$ from the Q-network for every $N^{\boldsymbol{\theta}}$ episodes. By selecting a batch of samples from the replay memory instead of a single sample, using two networks with the same structure but different parameters (the Q-network and the target Q-network), and updating the parameter vector $\boldsymbol{\theta}^-$ periodically, the stability of the learning process is improved.
\begin{algorithm}[h]
\DontPrintSemicolon
    \SetAlgoLined
    \KwIn{Action space $\mathcal{A}$, $N^\mathrm{I}$,  $N^{\boldsymbol{\theta}}$, $\lambda_{\mathrm{RMS}}$, $\gamma $, and $\epsilon$.}
    Initialization of replay memory $M$, $\boldsymbol{\theta}$, and $\boldsymbol{\theta}^-$.\\
    \For{$Iteration\leftarrow1$ to $N^\mathrm{I}$}
    {
        \For{$i\leftarrow1$ to $N^\mathrm{TTI}$}
        {
            Update the wireless channel model.\\
            \eIf {the probability is less than $\epsilon$}
            {
                Randomly select the action $A_i \in \mathcal{A}$.\\
            }
            {
                Select the action $A_i=\arg\max_{A}Q\left(S_i,A;\boldsymbol{\theta}_i\right)$.\\
            }
            The BS observes the state $S_{i+1}$ and calculate the reward of the action $R_i$.\\
            Store the transition $\left(S_i,A_i,R_i,S_{i+1}\right)$ in the replay memory $M$.\\
            Randomly sample transitions from replay memory $M$ with the number of batch size.\\
            Calculate gradient descent using Eq. (12).\\
            Update $\boldsymbol{\theta}$ using Eq. (13).\\
        }
        Update $\boldsymbol{\theta^-=\theta}$ in every $N^{\boldsymbol{\theta}}$ episodes.\\
    }
    \caption{DQN-based C\&C Data Generation}
\end{algorithm}
\par To balance the relationship between exploration and exploitation, we adopt the $\epsilon -$greedy approach in our learning process where $\epsilon \in \left[0,1\right]$. In each TTI, the agent randomly generates a probability and compares it with $\epsilon$. If the probability is less than $\epsilon$, the agent randomly chooses an action in this TTI. If not, the agent chooses the optimal action.
\par Based on that, we implement the DQN algorithm, which is presented in $\mathbf{Algorithm\ 2}$. At the beginning of $i^\mathrm{th}$ TTI, the agent at the BS selects an action $A_i$ which is utilized to form the C\&C data $\boldsymbol{m}_i$, and the BS transmit $\boldsymbol{m}_i$ to the UAV. At the end of this TTI $t_{i}$, the UAV sends its position to the BS. Then, the agent can obtain the next state $S_{i+1}$ and reward $R_i$. As the algorithm iterates, the parameters can be updated and optimal actions can be selected.

\subsubsection{Proactive Repetition Scheme} 
Due to the unstable communication channel, the packet loss and delay of downlink C\&C data transmission can be harmful to the task accomplishment. In order to increase the successful transmission of the C\&C data, we introduce the proactive scheme proposed in uplink transmission and been proven to have the lowest latent access failure probability (the failure probability within a limited time) among the reactive scheme, K-repetition scheme and itself \cite{Proactive}. In the proactive scheme for uplink transmission, the user equipment (UE) repeats the transmission for a maximum number of $K_\mathrm{max}$, where the period to send each repetition is denoted as $T_\mathrm{rep}$. The UE is configured to repeat the transmission for $K_\mathrm{max}$ repetitions but can receive feedback after each repetition. This allows the UE to terminate repetitions earlier once receiving positive feedback (ACK).
\par
\begin{figure}[htp]
    \centering
    \includegraphics[width=0.8\linewidth]{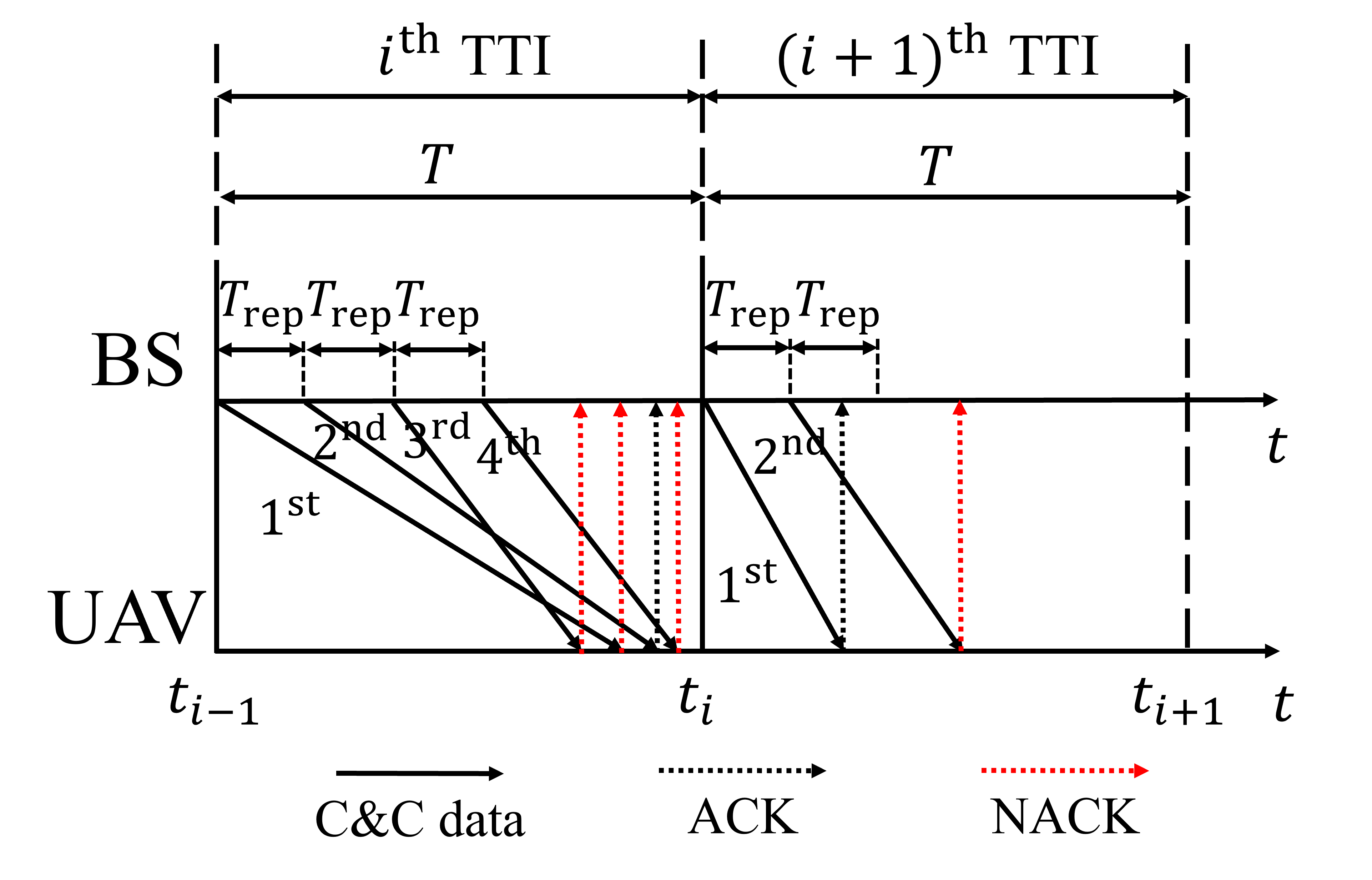}
    \caption{An example of the timeline of UAV and BS}
    \label{Timeline_Pro}
\end{figure}

 Different from the uplink transmission without the latency constraint, in our work, we focus on downlink transmission with stringent delay limitations. Thus, we do not take retransmission into consideration. The exemplar timeline between the BS and the UAV for the downlink C\&C data transmission is shown in Fig. \ref{Timeline_Pro}.
We make the assumption that the data (i.e., C\&C data, ACK, etc.) processing time is ignored. Then, the important parameters of the proactive repetition scheme are $K_\mathrm{max}$ and $T_\mathrm{rep}$. In this example, $K_\mathrm{max}$ is set to be 4. In the $i^\mathrm{th}$ TTI, only the $2^\mathrm{nd}$ repetition is successfully received by the UAV and its corresponding ACK arrival time is larger than the transmitting time of the $4^\mathrm{th}$ repetition. As a result, there is no repetition terminated at the BS and the number of repetitions is $K_\mathrm{max}$. By contrast, in the $\left(i+1\right)^\mathrm{th}$ TTI, the ACK of the $1^\mathrm{st}$ repetition is successfully received before the transmitting time of the $3^\mathrm{rd}$ repetition. Hence, the $3^\mathrm{rd}$ repetition is terminated and there are only two repetitions. It is worth noticing that the time duration between the first repetition and the maximum repetition should not exceed the threshold which is set to be the length of one TTI $T$.
\begin{algorithm}[h]
    \caption{Proactive Repetition Scheme}
    \KwIn{$K_\mathrm{max}$, $T_\mathrm{rep}$.} 
    Obtain the C\&C data $\boldsymbol{m}_i$.\\
    Initialize the arrival time $t_i^\mathrm{A}$ of $\boldsymbol{m}_i$ to $t_{i}$.\\
    \For{$k \leftarrow 1$ to $K_\mathrm{max}$}
    {
        \If{$t_i^\mathrm{A} \leq t_{i,k}=t_{i-1}+\left(k-1\right)T_\mathrm{rep}$}
        {
            break.\\
        }
        Update the UAV's position $\boldsymbol{p}_{i,k}$ at $t_{i,k}$.\\
        Calculate the transmission result $\delta_{i,k}^\mathrm{CC}$ and transmission time $t_{i,k}^\mathrm{CC}$ of $k^\mathrm{th}$ repetition.\\
        \If{$\delta_{i,k}^\mathrm{CC}=1$.}
        {
            \If{there is no $\boldsymbol{m}_i$ in $\mathcal{C}$.}
            {
                Add $\boldsymbol{m}_i$ in $\mathcal{C}$.\\
            }
        }
        \If{$t_i^\mathrm{A}>t_{i,k}+t_{i,k}^\mathrm{CC}$.}
        {
            $t_i^\mathrm{A}=t_{i,k}+t_{i,k}^\mathrm{CC}$.\\
        }
        \eIf{$k=K_\mathrm{max}$}
            {
                $t^\mathrm{End}=t_{i}$.\\
            }
            {
                $t^\mathrm{End}=t_{i,\left(k+1\right)}$.\\
            }
        Initialize $\mathcal{C}^\mathrm{Tmp}$ to $\emptyset$.\\
        \For{C\&C data in $\mathcal{C}$}
            {
                \If{this C\&C data arrives in $\left[t_{i,k},t^\mathrm{End}\right)$}
                {
                    Add this C\&C data to $\mathcal{C}^\mathrm{Tmp}$.\\
                }
            }
                
        Reorder the C\&C data in $\mathcal{C}^\mathrm{Tmp}$ by their arrival time.\\
        \For{all C\&C data in $\mathcal{C}^\mathrm{Tmp}$}
        {
            Update the UAV's position.\\
        }
    }
    Initialize $\mathcal{C}^\mathrm{Tmp}$ to $\emptyset$.\\
    \For{C\&C data in $\mathcal{C}$}
    {
        \If{the arrived time for the C\&C data is in $\left[t^\mathrm{End},t_{i}\right)$}
        {
            Add this C\&C data in $\mathcal{C}^\mathrm{Tmp}$.\\
        }
    }
    Reorder the C\&C data in $\mathcal{C}^\mathrm{Tmp}$ by their arrival time.\\
    \For{all C\&C data in $\mathcal{C}^\mathrm{Tmp}$}
    {
        Update the UAV's position.\\
    }
\end{algorithm}
\par The implementation of the proactive repetition scheme is presented in $\mathbf{Algorithm\ 3}$. After one C\&C data $\boldsymbol{m}_i$ is generated, its $k^\mathrm{th}$ ($k\in\left\{1,...,K_\mathrm{max}\right\}$) repetition occurs at $t_{i,k}=t_{i-1}+\left(k-1\right)T_\mathrm{rep}$. In the time duration between two consecutive repetitions, the UAV's position may change, which may influence the transmission process (i.e., transmission result $\delta^\mathrm{CC}$ and transmission time $t^\mathrm{CC}$) of future repetitions. As a result, we need to calculate the UAV's position in the time duration $\left[t_{i,k},t^\mathrm{End}\right)$, where $t^\mathrm{End}$ can be expressed as
\begin{equation}
    t^\mathrm{End}=\begin{cases}
        t_{i,\left( k+1 \right)},&\ k<K_\mathrm{max}\\
        t_{i},&\ k=K_\mathrm{max}.
    \end{cases}
\end{equation}
If $k=K_\mathrm{max}$, this means there is no future repetition of $\boldsymbol{m}_i$, the end of this time duration $t^\mathrm{End}$ is the beginning of the next TTI $t_{i}$. Otherwise, $t^\mathrm{End}$ is the transmission time of the next potential repetition $t_{i,\left(k+1\right)}$.
\par Since the queue at the UAV only stores the earliest successfully decoded C\&C data $\boldsymbol{m}_i$ for $i^\mathrm{th}$ TTI despite the number of successful repetitions it receives, we also need to obtain the smallest arrival time of all successfully decoded repetitions and consider it as the arrival time $t_i^\mathrm{A}$ of $\boldsymbol{m}_i$. Based on the arrival time of all C\&C data in the successfully decoded C\&C data set $\mathcal{C}$, we can obtain the C\&C data arrived during $\left[t_{i,k},t^\mathrm{End}\right)$ and update the UAV's position. This loop continues to iterate until the number of repetitions reaches $K_\mathrm{max}$ or the repetition is terminated when the ACK of the earliest successfully decoded repetition arrives at the BS ($t_i^\mathrm{A}<t_{i,k}$).

\subsubsection{DeepPro}
The DeepPro at the BS is designed by integrating the proactive repetition scheme into DRL-based C\&C data generation. In $i^\mathrm{th}$ TTI, the DRL-based C\&C data generator generates a C\&C data $\boldsymbol{m}_i$ based on $\mathbf{Algorithm\ 2}$, and the proactive repetition scheme repeats the transmission of $\boldsymbol{m}_i$ several times and the UAV's position is updated based on $\mathbf{Algorithm\ 3}$. Next, the UAV's position $\boldsymbol{p}_{i}$ at the end of the $i^\mathrm{th}$ TTI $t_{i}$ is obtained and utilized to acquire $S_{i+1}$ and $R_i$. Finally, the sample $\left(S_i,A_i,S_{i+1},R_i\right)$ is fed in the DQN training process in $\mathbf{Algorithm\ 2}$.

\subsection{VA-QOM} With the aim of executing the optimal C\&C data in the queue to accomplish the task, we design a VA-QOM scheme to control the UAV to fly to the target position as close as possible, where the size of the queue is set to be $Q_\mathrm{max}$. Different from the traditional UAV control framework where the queue is ordered by the arrival time, we rank the queue according to semantic information, which is described by the AoI and the VoI.

\subsubsection{AoI Design} Since the packet loss and transmission delay may cause negative impacts (i.e., delayed response time and wrong actions of the UAV) on the accomplishment of our time-sensitive task, it is crucial to utilize the AoI to evaluate the freshness of the packet. In our task, the AoI of the C\&C data $\boldsymbol{m}_i$ can be formulated as $\mathrm{AoI}\left(\boldsymbol{m}_i\right)=t-t_{i-1}$, where $t$ is the current time and $t_{i-1}$ is the $\boldsymbol{m}_i$ generation time instance. The higher the AoI is, the less fresh the C\&C data is.

\subsubsection{VoI Design} In our task, we aim to decrease the difference between the UAV's actual trajectory and its target trajectory, which converts to decreasing the distance between the UAV's actual positions and the UAV's target positions at each TTI. It is important to note that different C\&C data may have different importance or value to the task, and the value of the same C\&C data may change according to time. Concretely, at the same time, different C\&C data may control the UAV to move closer or further to its target position, which implies different values. For the same C\&C data, even though the content of data (i.e., planned velocity) is the same, the target position of the UAV may vary from time due to dynamics during transmission. To quantify the value of the C\&C data in the current time $t$, we introduce the VoI, denoted as $\mathrm{VoI}\left(\boldsymbol{m}_i\right)$.
\par We assume that when the UAV ranks the queue and executes the C\&C data according to the order at $t$, there is $n$ ($n \in \mathbbm{N}$) C\&C data in the queue $ \left\{\boldsymbol{m}^{\mathrm{Q}_1},...,\boldsymbol{m}^{\mathrm{Q}_l},...\boldsymbol{m}^{\mathrm{Q}_n}\right\}$ $\left(l\in \left\{1,...,n\right \}\right)$. Based on the assumption, for any queue updating time $t$, the index of TTI of this queue updating time $t$ is written as
\begin{equation}
    i^\mathrm{E}=\lfloor \frac{t}{T}\rfloor+1,t\in\left[ 0,TN^\mathrm{TTI} \right)
\end{equation}
where $\lfloor . \rfloor$ is the floor function. Once the queue is updated, the first C\&C data in the queue will be executed. 
\par At time $t\in \left[t_{i^\mathrm{E}-1},t_{i^\mathrm{E}}\right)$, the UAV aims to fly to the its target position $\boldsymbol{g}_{i^\mathrm{E}}$ as close as possible at $t_{i^\mathrm{E}}$. As the UAV has no direct knowledge about its target position $\boldsymbol{g}_{i^\mathrm{E}}$ and cannot know its actual future position $\boldsymbol{p}_{i^\mathrm{E}}$, we propose a method for the UAV to estimate them. The estimated target position of the UAV at $t_{i^\mathrm{E}}$ is denoted as $\hat{\boldsymbol{g}}_{i^\mathrm{E}}$. For the estimated actual position of the UAV at $t_{i^\mathrm{E}}$, due to that different C\&C data $\boldsymbol{m}^{\mathrm{Q}_l}$ in the queue may control the UAV to fly to different positions, different C\&C data $\boldsymbol{m}^{\mathrm{Q}_l}$ has different UAV's estimated actual position $\hat{\boldsymbol{p}}_{i^\mathrm{E},\mathrm{Q}_l}$. For the C\&C data $\boldsymbol{m}^{\mathrm{Q}_l}$ in the queue, if its estimated actual position $\hat{\boldsymbol{p}}_{i^\mathrm{E},\mathrm{Q}_l}$ is closer to the estimated target position $\hat{\boldsymbol{g}}_{i^\mathrm{E}}$, it will be more valuable. As a result, we define the VoI of $\boldsymbol{m}^{\mathrm{Q}_l}$ as
\begin{equation}
    \mathrm{VoI}\left( \boldsymbol{m}^{\mathrm{Q}_l} \right)=-||\hat{\boldsymbol{p}}_{i^\mathrm{E},\mathrm{Q}_l}-\hat{\boldsymbol{g}}_{i^\mathrm{E}}||,
\end{equation}
where it is the negative value of the distance between the estimated actual position and estimated target position at $t_{i^\mathrm{E}}$. The lower it is, the more valuable this C\&C data is.
\par
\begin{figure}[!t]
    \centering
    \includegraphics[width=0.65\linewidth]{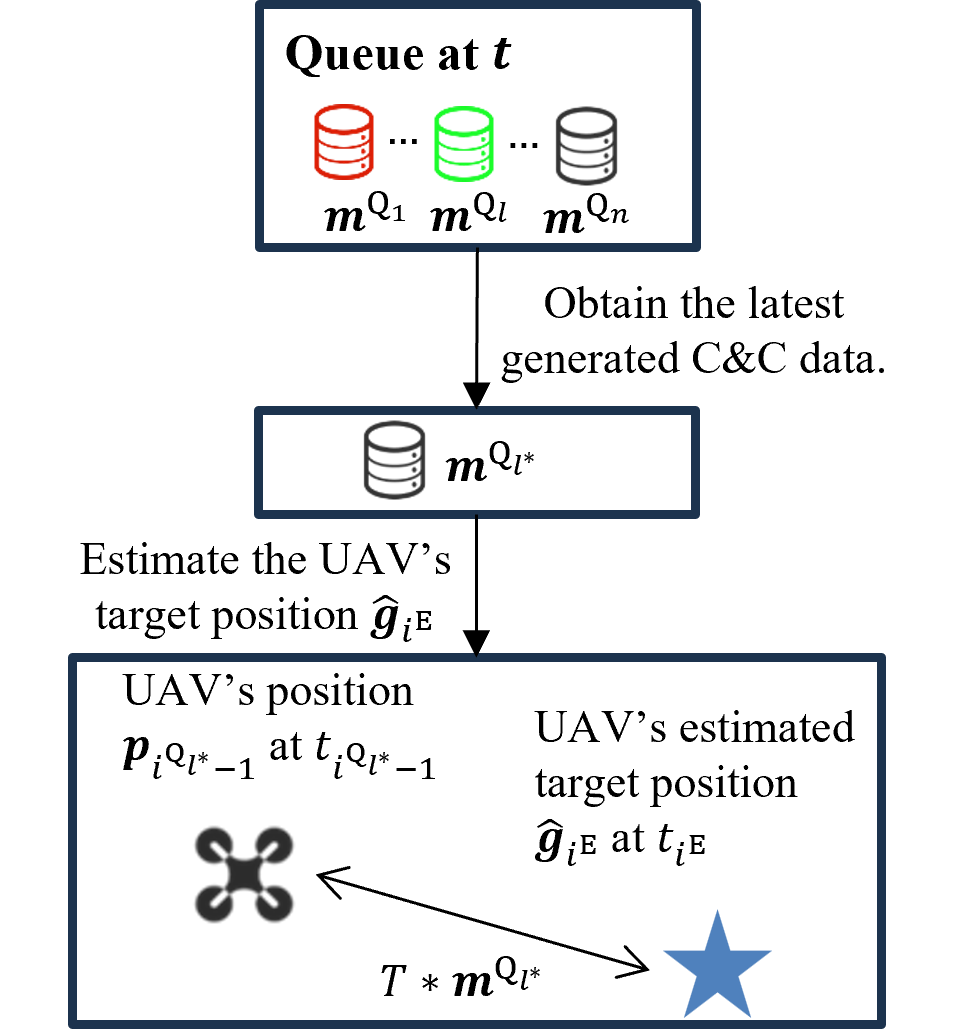}
    \caption{The process of estimating the UAV's target position $\hat{\boldsymbol{g}}_{i^\mathrm{E}}$.}
    \label{Target position estimation}
\end{figure}

 Next, we present how we obtain $\hat{\boldsymbol{g}}_{i^\mathrm{E}}$ and $\hat{\boldsymbol{p}}_{i^\mathrm{E},\mathrm{Q}_l}$, respectively. We define $\hat{\boldsymbol{g}}_{i^\mathrm{E}}$ as the latest position where the BS expects the UAV to reach, which can be obtained in two steps, as shown in Fig. \ref{Target position estimation}.
\begin{figure}[!t]
    \centering
    \includegraphics[width=0.65\linewidth]{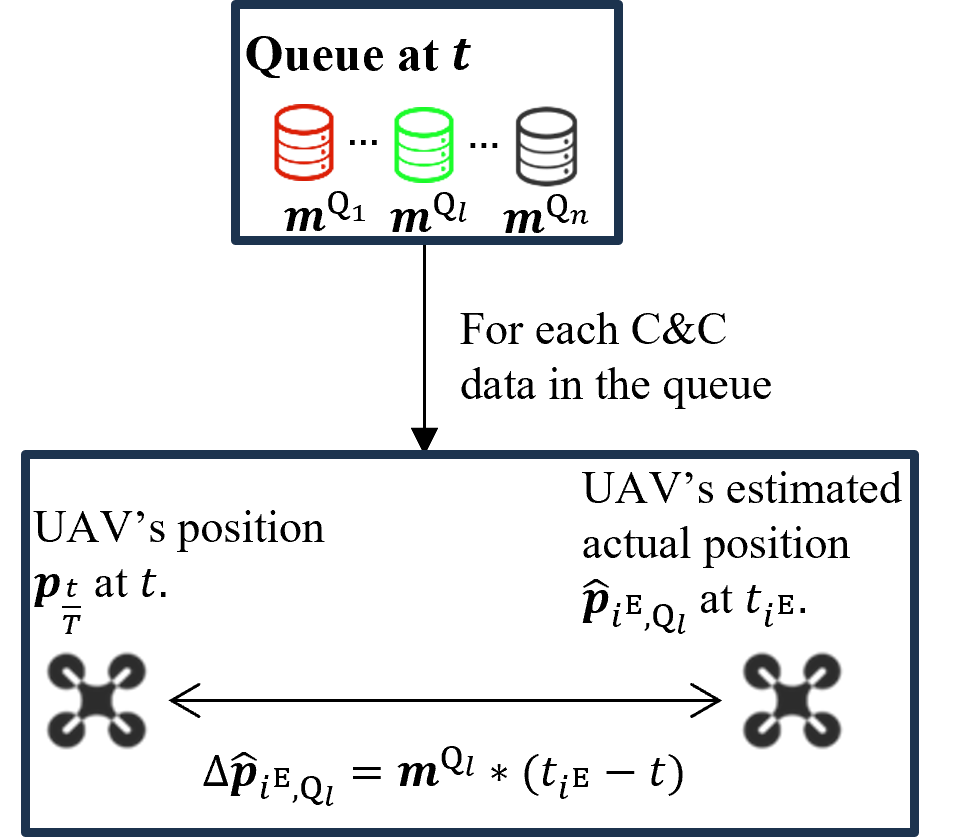}
    \caption{The process of estimating the UAV's actual position $\hat{\boldsymbol{p}}_{i^\mathrm{E},\mathrm{Q}_l}$.}
    \label{Actual position estimation}
\end{figure}
First, we calculate the AoI of all C\&C data $\boldsymbol{m}^{\mathrm{Q}_{l}}$ in the queue to find the latest generated C\&C data $\boldsymbol{m}^{\mathrm{Q}_{l^*}}$. For $l^\mathrm{th}$ C\&C data $\boldsymbol{m}^{\mathrm{Q}_{l}}$, its AoI is calculated as
\begin{equation}
\mathrm{AoI}\left(\boldsymbol{m}^{\mathrm{Q}_{l}}\right)=t-t_{i^{\mathrm{Q}_l}-1},
\end{equation}
where $t_{i^{\mathrm{Q}_l}-1}$ is the generation time of $\boldsymbol{m}^{\mathrm{Q}_{l}}$. Based on that, we can obtain the index of the C\&C data in the queue with the smallest AoI as
\begin{equation}
l^*=\arg\min_{l}\mathrm{AoI}\left(\boldsymbol{m}^{\mathrm{Q}_{l}}\right).
\end{equation}
Second, due to the C\&C data $\boldsymbol{m}^{\mathrm{Q}_{l^*}}$ is the planned velocity vector $\left( v_{i^{\mathrm{Q}_{l^*}}}^x, v_{i^{\mathrm{Q}_{l^*}}}^y, z_{i^{\mathrm{Q}_{l^*}}}^z\right)$ of the UAV and its planned execution time is $T$, we are able to attain the planned movement of the UAV as $T\boldsymbol{m}^{\mathrm{Q}_{l^*}}$. Based on that, we estimate the target position for the UAV at time $t_{i^\mathrm{E}}$ as
\begin{equation}
\hat{\boldsymbol{g}}_{i^\mathrm{E}}=\boldsymbol{p}_{i^{\mathrm{Q}_{l^*}}-1}+T\boldsymbol{m}^{\mathrm{Q}_{l^*}},
\end{equation}
where $\boldsymbol{p}_{i^{\mathrm{Q}_{l^*}}-1}$ is the UAV's actual position at the generation time $t_{i^{\mathrm{Q}_{l^*}}-1}$ of the C\&C data $\boldsymbol{m}^{\mathrm{Q}_{l^*}}$ in the queue.
\par Then, we estimate the UAV's actual position $\hat{\boldsymbol{p}}_{i^\mathrm{E},\mathrm{Q}_l}$ for the C\&C data $\boldsymbol{m}^{\mathrm{Q}_l}$ in two steps, as shown in Fig. \ref{Actual position estimation}. First, as the queue is updated at $t$, the UAV will execute the C\&C data $\boldsymbol{m}^{\mathrm{Q}_l}$ from $t$ to $t_{i^\mathrm{E}}$, with the movement as
\begin{equation}
    \Delta \hat{\boldsymbol{p}}_{i^\mathrm{E},\mathrm{Q}_l}=\boldsymbol{m}^{\mathrm{Q}_l}\left(t_{i^\mathrm{E}}-t\right).
\end{equation}
Second, since the UAV know its current position $\boldsymbol{p}_{\frac{t}{T}}$ at $t$, we estimate the actual position of the UAV at time $t_{i^\mathrm{E}}$ as
\begin{equation}
\hat{\boldsymbol{p}}_{i^\mathrm{E},\mathrm{Q}_l}=\boldsymbol{p}_{\frac{t}{T}}+\Delta \hat{\boldsymbol{p}}_{i^\mathrm{E},\mathrm{Q}_l}.
\end{equation}
After estimating $\hat{\boldsymbol{g}}_{i^\mathrm{E}}$ and $\hat{\boldsymbol{p}}_{i^\mathrm{E},\mathrm{Q}_l}$, the VoI $\mathrm{VoI}\left(\boldsymbol{m}^{\mathrm{Q}_l}\right)$ of the C\&C data $\boldsymbol{m}^{\mathrm{Q}_l}$ can be calculated using Eq. (16).
\subsubsection{Semantic Information Extraction}
As the optimal C\&C data is generated at the beginning of each TTI, the C\&C data in the queue is most valuable for the UAV if its AoI is less than $T$. Otherwise, its semantic information only depends on its VoI. To normalize the range of semantic information, we define the function of semantic information for $\boldsymbol{m}^{\mathrm{Q}_l}$ as
\begin{multline}
    \mathrm{SI}\left(\boldsymbol{m}^{\mathrm{Q}_l}\right)=u\left[T-\mathrm{AoI}\left(\boldsymbol{m}^{\mathrm{Q}_l}\right)\right]\\
    +u\left[\mathrm{AoI}\left(\boldsymbol{m}^{\mathrm{Q}_l}\right)-T\right]e^{\mathrm{VoI}\left(\boldsymbol{m}^{\mathrm{Q}_l}\right)},
\end{multline}
where the function $u(.)$ is formulated as
\begin{equation}
    u\left(x\right)=\begin{cases}
        1,\quad x>0\\
        0,\quad x\leq0.
    \end{cases}
\end{equation}
If $\mathrm{AoI}\left(\boldsymbol{m}^{\mathrm{Q}_l}\right)<T$, $\boldsymbol{m}^{\mathrm{Q}_l}$ is just generated at the beginning of this TTI, thus it has the highest semantic information which is equal to 1. Otherwise, $\boldsymbol{m}^{\mathrm{Q}_l}$ is assumed to be outdated and its $\mathrm{VoI}\left(\boldsymbol{m}^{\mathrm{Q}_l}\right)$ decides its importance to the task. As the $\mathrm{VoI}\left(\boldsymbol{m}^{\mathrm{Q}_l}\right)$ increases, its semantic information increases and the upper bound is 1. At the beginning of $i^\mathrm{th}$ TTI $t_{i-1}$ or when the UAV successfully decodes one new C\&C data at $t_i^\mathrm{A}$, the UAV starts to order the queue based on the semantic information in a descending order, as shown in $\mathbf{Algorithm\ 4}$.  
\begin{algorithm}[htbp]
    \caption{VA-QOM}
    \KwIn{The UAV's queue $\mathcal{Q}$, at the beginning of $i^\mathrm{th}$ TTI $t=t_{i-1}$ or at the time $t=t_i^\mathrm{A}$ when the UAV successfully decodes a C\&C data.}  
    \KwOut{The UAV's reordered queue $\mathcal{Q}^\mathrm{R}$} 
    Calculate $i^\mathrm{E}$ using Eq. (15).\\
    \For{$l\leftarrow 1$ to $n$}
    {
        Calculate $\Delta\hat{\boldsymbol{p}}_{i^\mathrm{E},\mathrm{Q}_l}$ from $t$ to $t_{i^\mathrm{E}}$ using Eq. (20).\\
        Calculate $\hat{\boldsymbol{p}}_{i^\mathrm{E},\mathrm{Q}_l}$ at $t_{i^\mathrm{E}}$ using Eq. (21).\\
        Calculate $\mathrm{AoI}\left(\boldsymbol{m}^{\mathrm{Q}_{l}}\right)$ using Eq. (17).\\
    }
    Obtain the index $l^*$ using Eq. (18).\\
    Obtain $\hat{\boldsymbol{g}}_{i^\mathrm{E}}$ at $t_{i^\mathrm{E}}$ using Eq. (19).\\
    \For{$l\leftarrow 1$ to $n$}
    {
        Calculate $\mathrm{VoI}\left(\boldsymbol{m}^{\mathrm{Q}_{l}}\right)$ using Eq. (16).\\
        Calculate $\mathrm{SI}\left(\boldsymbol{m}^{\mathrm{Q}_{l}}\right)$ using Eq. (22).\\
    }
    Reorder the queue $\mathcal{Q}$ to $\mathcal{Q}^\mathrm{R}$ based on the semantic information in descending order.\\
\end{algorithm}
\subsection{Communication and Control Process}
\begin{figure}[!t]
    \centering
        \includegraphics[width=0.9\linewidth]{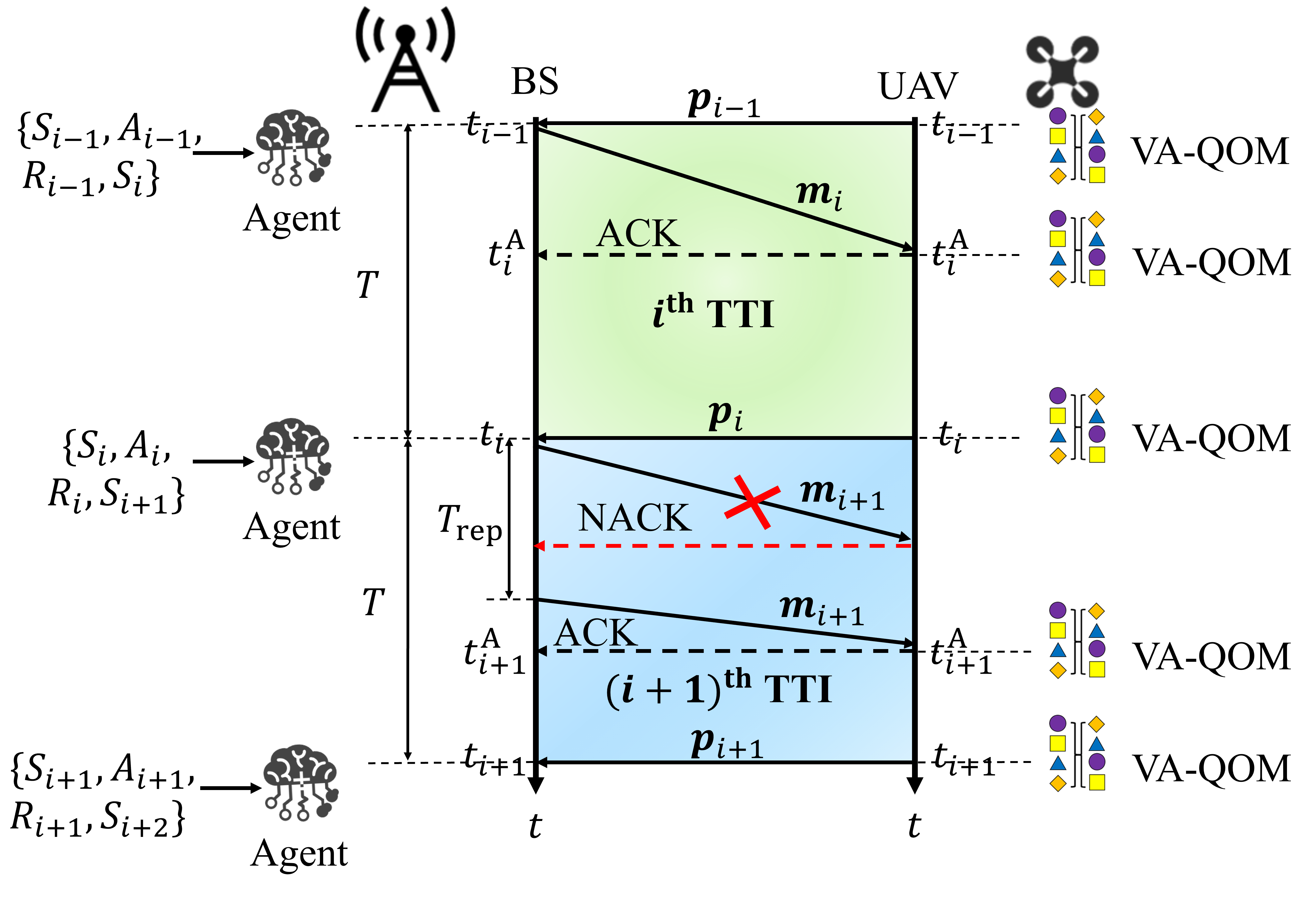}
        \caption{Exemplar timeline in our proposed GSRC framework.}
        \label{GSRC_timeline}
\end{figure}

As shown in Fig. \ref{GSRC_timeline}, we plot the exemplar timeline of the BS and the UAV to introduce the communication and control process in our proposed GSRC framework. At the beginning of $i^\mathrm{th}$ TTI $t_{i-1}$, the agent at the BS selects the action $A_i$ from past observations to generate the C\&C data $\boldsymbol{m}_i$ which is transmitted by the proactive repetition scheme. At $t_{i-1}$ or at $t_i^\mathrm{A}$ When the UAV successfully decodes this C\&C data $\boldsymbol{m}_i$, the UAV reorders its queue based on the semantic information and executes the C\&C data in the queue according to its order to update its position. At the end of $i^\mathrm{th}$ TTI $t_{i}$, the UAV sends its real-time position $\boldsymbol{p}_{i}$ back to the BS. The agent at the BS utilizes $\boldsymbol{p}_{i}$ to obtain the reward $R_i$ and the next state $S_{i+1}$ which are stored in the observation. This process continues looping until all the C\&C data are transmitted.

\section{Simulation Results}
In this section, we examine the effectiveness of our proposed GSRC framework and compare it with the traditional UAV control framework (TUCF) via simulation. To assess the individual contributions of the GSRC components, we also conduct separate simulations for the DeepPro and VA-QOM. The start position of the UAV is $\boldsymbol{p}_0=\boldsymbol{g}_0=\left(80\mathrm{m},80\mathrm{m},20\mathrm{m}\right)$ at time $t=0\mathrm{s}$. We set $T$ and $N^\mathrm{TTI}$ as 1 ms and 99, respectively. For parameters in the channel model, we adopt $f_c$, $\gamma_\mathrm{th}$, $\sigma^2$, and $\mathrm{P}$ as 5 GHz, 5.5 dB, -104 dBm, and 18 dBm, respectively. The hyperparameters $\epsilon$, $\gamma$, $\lambda_\mathrm{RMS}$ of the DQN algorithm are set to 1, 0.1, and $10^{-4}$, respectively. The size of one C\&C data is 104 bytes, and the queue size $Q_\mathrm{max}$ and the parameter $N^\mathrm{M}$ are 10 and 9, respectively. The velocity sets are $V^\mathrm{X},\ V^\mathrm{Y}\in\left\{-5000,-4000,...,5000\right\}$ and $V^\mathrm{Z}\in\left\{0\right\}$, respectively. The results are obtained by simulating each algorithm 1000 times and taking the average value.

\begin{figure*}[htbp]
    \centering
    \subfigure[Trajectory comparison.]{
    \includegraphics[width=0.3\textwidth]{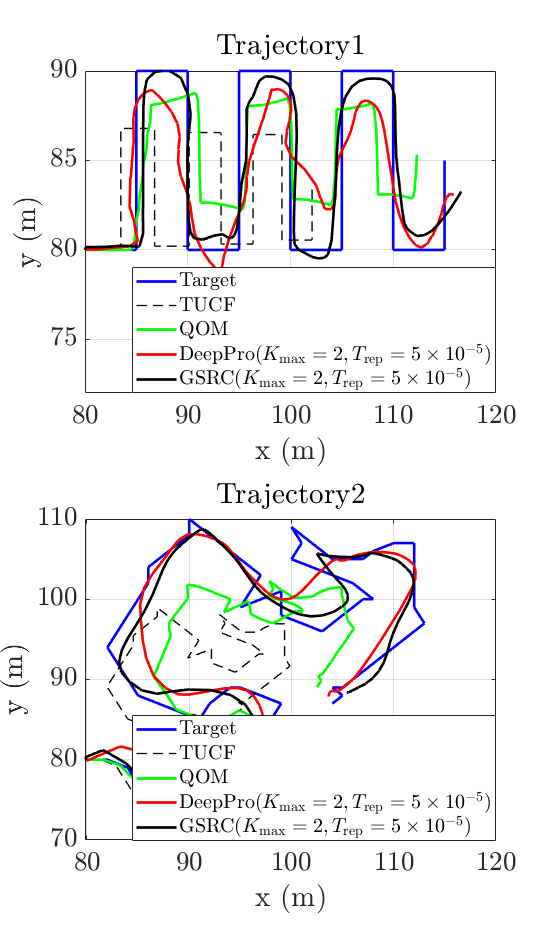}
    }
    \subfigure[Error comparison.]{
    \includegraphics[width=0.3\textwidth]{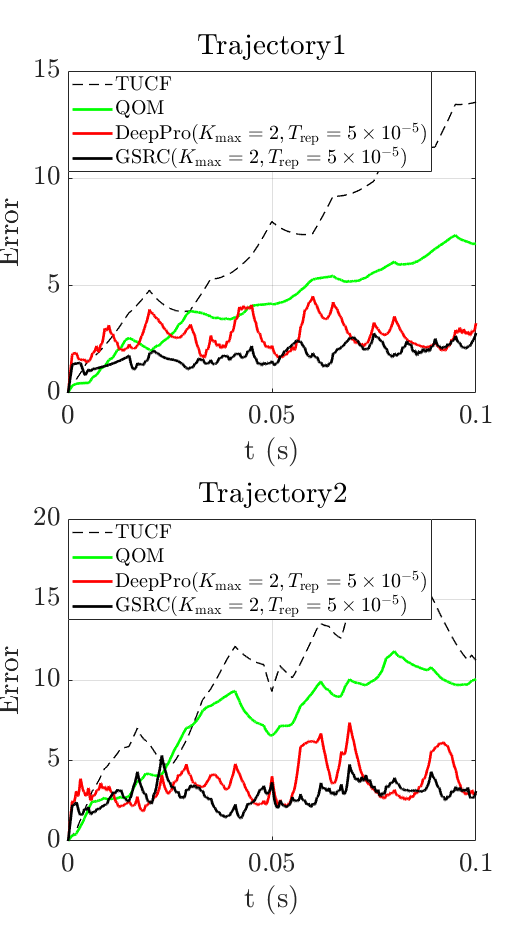}
    }
    \subfigure[MSE comparison.]{
    \includegraphics[width=0.3\textwidth]{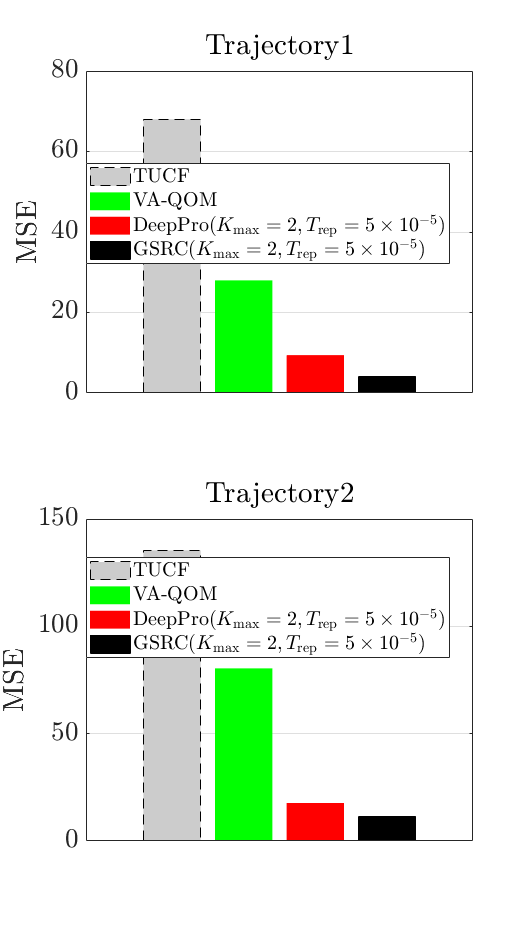}
    }
    \caption{The comparison of various schemes.}
    \label{Comparison}
\end{figure*}

\par We compare different schemes (TUCF, VA-QOM, DeepPro, and GSRC) via simulation using two types of trajectories, as shown in Fig. \ref{Comparison}. The first trajectory is a simple one but typical for our use case, while the second is more complex with a randomly generated trajectory. Fig. \ref{Comparison} (a) plots the trajectory comparison over different schemes. We can observe that the trajectory of the TUCF is notably most distant from the target trajectory. The TUCF and the VA-QOM share a similar trajectory shape because their generated C\&C data is the same. However, the VA-QOM, with its ability to update target positions and prioritise the queue, achieves a trajectory closer to the target one compared to the TUCF. Due to the severe packet loss and delay, the update in the VA-QOM is always not on time, which still results in a huge gap between the trajectory of the VA-QOM and the target one. The DeepPro's trajectory is closer to the target one than the VA-QOM because it can generate optimal C\&C data via the DQN algorithm and decrease the risk of packet loss by the proactive repetition scheme. By integrating the DeepPro and the VA-QOM, the trajectory of our proposed GSRC framework is the closest to the target trajectory, indicating superior performance.
\par Fig. \ref{Comparison} (b) plots the error comparison for different schemes over time. It is noticed that our proposed GSRC achieves the lowest overall error, with the DeepPro and VA-QOM also showing lower overall errors compared to TUCF. Interestingly, the error of the DeepPro and the GSRC remains stable over time, whereas the error of the TUCF and the VA-QOM fluctuate substantially over time. The reason is that the TUCF and VA-QOM are more sensitive to the communication environment with the same C\&C data sequence, while the DeepPro and the GSRC can dynamically adapt their generated C\&C data to the communication environment automatically. Fig. \ref{Comparison} (c) presents the MSE comparison over different schemes. The TUCF achieves the highest MSE. Compared to TUCF, both DeepPro and VA-QOM substantially reduce MSE by 86.92$\%$ and 40.61$\%$, respectively. The GSRC, based on the integration of DeepPro and VA-QOM, achieves the lowest MSE, resulting in a remarkable 91.51$\%$ reduction compared to TUCF.

\par
\begin{figure}[htbp]
    \centering
    \subfigure[$T_\mathrm{rep}=5 \times 10^{-5}$ (s).]{\includegraphics[width=0.65\linewidth]{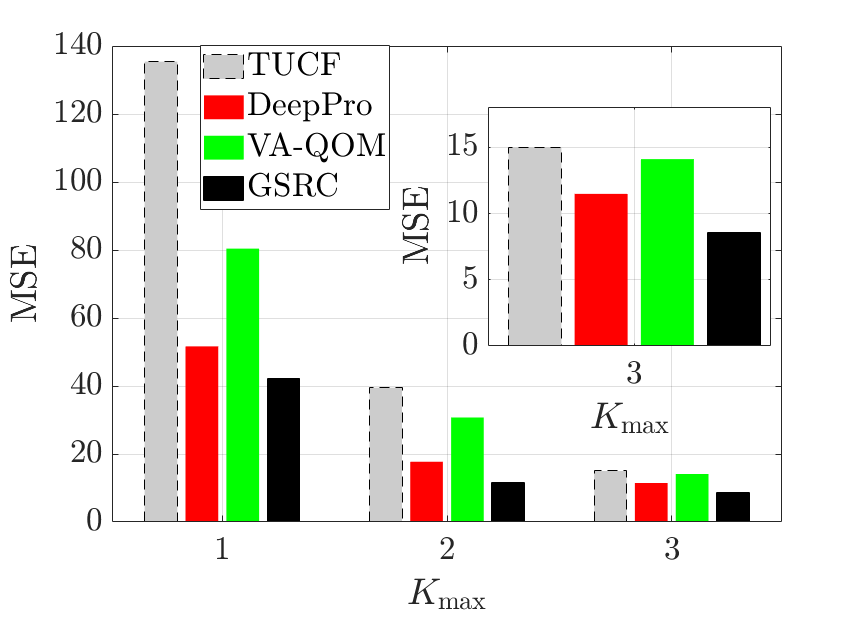}}
    \subfigure[$K_\mathrm{max}=3$.]{\includegraphics[width=0.65\linewidth]{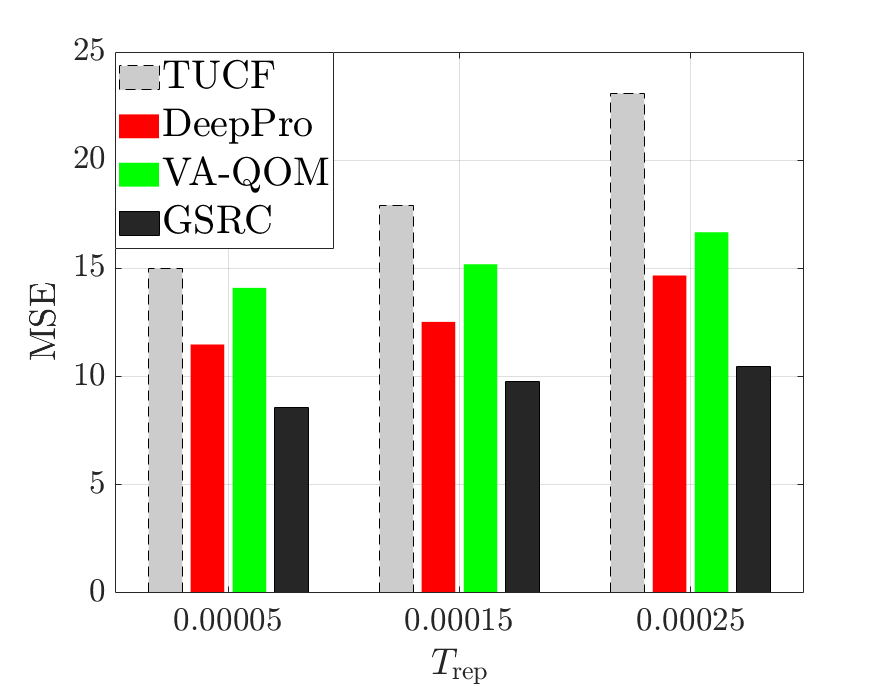}}    
    \caption{The MSE comparison over different values of $K_\mathrm{max}$ and $T_\mathrm{rep}$.}
    \label{MSE_K_T}
\end{figure}

 Fig. \ref{MSE_K_T} presents the impact of the proactive repetition scheme on the MSE for different schemes (TUCF, DeepPro, VA-QOM and GSRC). 
It is noticeable that our proposed GSRC scheme achieves the lowest MSE in various values of $K_\mathrm{max}$ and $T_\mathrm{rep}$, while the TUCF scheme has the highest MSE. Interestingly, the changes in $K_\mathrm{max}$ and $T_\mathrm{rep}$, which are highly related to the packet loss and delay of the C\&C data, have a more obvious impact on the MSE of the TUCF and VA-QOM schemes compared to the DeepPro and GSRC. Since the DeepPro and GSRC are able to generate optimal C\&C data to adapt to varying packet loss and delay conditions, their MSE remains more stable during such changes. Fig. \ref{MSE_K_T} (a) plots the MSE for various schemes over different values of $K_\mathrm{max}$ with a fixed $T_\mathrm{rep}=5\times10^{-5}$. We can observe that increasing $K_\mathrm{max}$ leads to a decrease in MSE for all schemes. This reduction is due to a decrease in packet loss of each C\&C data as $K_\mathrm{max}$ increases. Additionally, the MSE gap between the TUCF and DeepPro schemes diminishes because DQN is more likely to select the C\&C data which is the same as the TUCF when $K_\mathrm{max}$ is higher. Fig. \ref{MSE_K_T} (b) shows the MSE for various schemes against different values of $T_\mathrm{rep}$ with a fixed $K_\mathrm{max}=3$. It can be observed that with the $T_\mathrm{rep}$ increasing, there is a rise in the MSE of all schemes. The reason is that with a larger $T_\mathrm{rep}$, the BS needs more time to transmit the next repetition if there is no ACK back, which increases the average arrival time of each C\&C data.
\begin{figure}[htbp]
    \centering
    \includegraphics[width=0.7\linewidth]{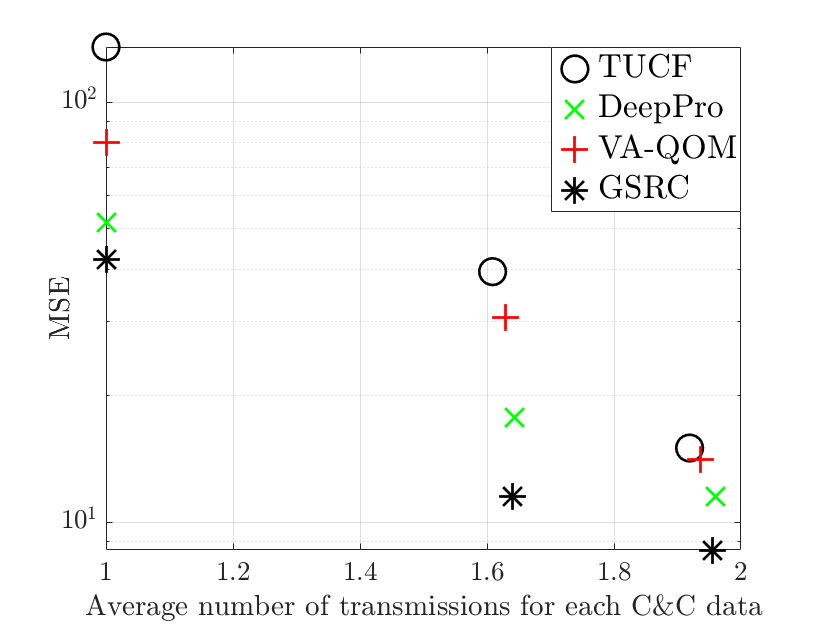}
    \caption{The relationship between the MSE and the average number of transmissions for each C\&C data when $T_\mathrm{rep}=5\times10^{-5}$.}
    \label{Ave_MSE}
\end{figure}
\par Fig. \ref{Ave_MSE} plots the relationship between the MSE and the average number of actual transmissions for each C\&C data over time of all schemes when $T_\mathrm{rep}=5\times10^{-5}$ with various $K_\mathrm{max}$. We can observe that as the average number of transmissions per C\&C data increases, the MSE for all schemes decreases. Among all schemes, our proposed GSRC framework maintains the lowest MSE, while the TUCF exhibits the highest MSE when the average number of transmissions is nearly equivalent. This indicates that the design of the DeepPro and VA-QOM is able to enhance task performance, and the combination of them (GSRC) could significantly improve task performance while maintaining a similar number of transmissions. Moreover, we can see that our proposed GSRC framework requires fewer transmissions than the TUCF while achieving a comparable MSE, demonstrating its potential to reduce the number of transmissions while adhering to task performance constraints compared to the TUCF.

\section{Conclusion}
In this paper, we proposed a GSRC framework which incorporates both semantic-level and effectiveness-level for a real-time UAV waypoint transmission task. At the semantic level, we proposed the VA-QOM scheme at the UAV to rank the importance of data packets in the queue in accomplishing a task based on the AoI and the newly defined VoI. At the effectiveness level, we proposed a DeepPro scheme at the BS, which facilitates the generation of optimal C\&C data to accomplish the task and elevates the probability of successful transmission for each C\&C data by optimizing the MSE. Our simulation results shed light on that our proposed GSRC framework can decrease the number of average C\&C data transmissions when maintaining a similar MSE and enhance the MSE when maintaining a similar number of C\&C data transmissions.


%

\ifCLASSOPTIONcaptionsoff
  \newpage
\fi



%

\bibliographystyle{ieeetr}
\bibliography{mylib}

\begin{thebibliography}{10}

\bibitem{Robotics1}
Y.~Liu, X.~Ma, L.~Shu, G.~P. Hancke, and A.~M. Abu-Mahfouz, ``From industry 4.0 to agriculture 4.0: Current status, enabling technologies, and research challenges,'' {\em IEEE Trans. Ind. Informatics}, vol.~17, pp.~4322--4334, Jun. 2021.

\bibitem{Robotics2}
A.~Alam, ``Educational robotics and computer programming in early childhood education: A conceptual framework for assessing elementary school students’ computational thinking for designing powerful educational scenarios,'' in {\em Proc. 2022 Int. Conf. Smart Technologies Syst. Next Gener. Comput. (ICSTSN)}, pp.~1--7, Mar. 2022.

\bibitem{Robotics3}
T.~Haidegger, ``Autonomy for surgical robots: Concepts and paradigms,'' {\em IEEE Trans. Med. Robot. Bionics}, vol.~1, pp.~65--76, Apr. 2019.

\bibitem{Wireless_communication1}
S.~H. Alsamhi, O.~Ma, and M.~S. Ansari, ``Convergence of machine learning and robotics communication in collaborative assembly: Mobility, connectivity and future perspectives,'' {\em J. Intell. Robot. Syst.}, vol.~98, pp.~541--566, Oct. 2020.

\bibitem{Wireless_communication2}
Y.~Zeng, R.~Zhang, and T.~J. Lim, ``Wireless communications with unmanned aerial vehicles: Opportunities and challenges,'' {\em IEEE Commun. Mag.}, vol.~54, pp.~36--42, May 2016.

\bibitem{5G1}
M.~Shafi, A.~F. Molisch, P.~J. Smith, T.~Haustein, P.~Zhu, P.~De~Silva, F.~Tufvesson, A.~Benjebbour, and G.~Wunder, ``5{G}: A tutorial overview of standards, trials, challenges, deployment, and practice,'' {\em IEEE J. Sel. Areas Commun.}, vol.~35, pp.~1201--1221, Apr. 2017.

\bibitem{URLLC1}
H.~Ren, C.~Pan, Y.~Deng, M.~Elkashlan, and A.~Nallanathan, ``Joint power and blocklength optimization for {URLLC} in a factory automation scenario,'' {\em IEEE Trans. Wireless Commun.}, vol.~19, pp.~1786--1801, Dec. 2020.

\bibitem{URLLC2}
C.~Pan, H.~Ren, Y.~Deng, M.~Elkashlan, and A.~Nallanathan, ``Joint blocklength and location optimization for {URLLC}-enabled {UAV} relay systems,'' {\em IEEE Commun. Lett.}, vol.~23, pp.~498--501, Jan. 2019.

\bibitem{URLLC3}
B.~Chang, L.~Zhang, L.~Li, G.~Zhao, and Z.~Chen, ``Optimizing resource allocation in {URLLC for} real-time wireless control systems,'' {\em IEEE Trans. Veh. Technol.}, vol.~68, pp.~8916--8927, Jul. 2019.

\bibitem{5G2}
``Minimum requirements related to technical performance for {IMT}-2020 radio interface(s),'' tech. rep., doc. ITU-R M, Oct. 2016.

\bibitem{Drawback1}
C.~D. Alwis, A.~Kalla, Q.-V. Pham, P.~Kumar, K.~Dev, W.-J. Hwang, and M.~Liyanage, ``Survey on 6{G} frontiers: Trends, applications, requirements, technologies and future research,'' {\em IEEE Open J. Commun. Soc.}, vol.~2, pp.~836--886, Apr. 2021.

\bibitem{Drawback2}
J.~Morales, G.~Rodriguez, G.~Huang, and D.~Akopian, ``Toward {UAV} control via cellular networks: Delay profiles, delay modeling, and a case study within the 5-mile range,'' {\em IEEE Trans. Aerosp. Electron. Syst.}, vol.~56, pp.~4132--4151, Apr. 2020.

\bibitem{6G1}
G.~Gui, M.~Liu, F.~Tang, N.~Kato, and F.~Adachi, ``6{G}: Opening new horizons for integration of comfort, security, and intelligence,'' {\em IEEE Wireless Commun.}, vol.~27, pp.~126--132, Mar. 2020.

\bibitem{6G2}
``Framework and overall objectives of the future development of {IMT} for 2030 and beyond,'' tech. rep., ITU-T Tech. Rep., Jun. 2023.

\bibitem{6G_communication2}
A.~Ranjha, G.~Kaddoum, and K.~Dev, ``Facilitating {URLLC} in {UAV}-assisted relay systems with multiple-mobile robots for 6{G} networks: A prospective of agriculture 4.0,'' {\em IEEE Trans. Ind. Informatics}, vol.~18, pp.~4954--4965, Nov. 2022.

\bibitem{6G_communication1}
S.~Zhang, H.~Zhang, and L.~Song, ``Beyond {D2D}: Full dimension {UAV}-to-everything communications in 6{G},'' {\em IEEE Trans. Veh. Technol.}, vol.~69, pp.~6592--6602, Apr. 2020.

\bibitem{6G_communication3}
W.~Sun, H.~Zhang, R.~Wang, and Y.~Zhang, ``Reducing offloading latency for digital twin edge networks in {6G},'' {\em IEEE Trans. Veh. Technol.}, vol.~69, pp.~12240--12251, Aug. 2020.

\bibitem{Rob_Control1}
T.~Johannink, S.~Bahl, A.~Nair, J.~Luo, A.~Kumar, M.~Loskyll, J.~A. Ojea, E.~Solowjow, and S.~Levine, ``Residual reinforcement learning for robot control,'' in {\em Proc. 2019 Int. Conf. Robotics Autom. (ICRA)}, pp.~6023--6029, May 2019.

\bibitem{Rob_Control2}
C.~Yang, D.~Huang, W.~He, and L.~Cheng, ``Neural control of robot manipulators with trajectory tracking constraints and input saturation,'' {\em IEEE Trans. Neural Netw. Learn. Syst.}, vol.~32, pp.~4231--4242, Aug. 2021.

\bibitem{Transmitter}
P.~Iyer, S.~Tarekar, and S.~Dixit, ``Hand gesture controlled robot,'' in {\em Proc. 2019 9th Int. Conf. Emerging Trends Eng. Technol. - Signal Inf. Process. (ICETET-SIP-19)}, pp.~1--5, Nov. 2019.

\bibitem{TOSA1}
H.~{Zhou}, X.~{Liu}, Y.~{Deng}, N.~{Pappas}, and A.~{Nallanathan}, ``Task-oriented and semantics-aware {6G} networks,'' {\em arXiv e-prints}, p.~arXiv:2210.09372, Oct. 2022.

\bibitem{task_per1}
M.~Nasr-Azadani, J.~Abouei, and K.~N. Plataniotis, ``Single- and multiagent actor-critic for initial {UAV}’s deployment and {3-D} trajectory design,'' {\em IEEE Internet Things J.}, vol.~9, pp.~15372--15389, Feb. 2022.

\bibitem{task_per2}
Z.~Wang, R.~Zhou, C.~Hu, and Y.~Zhu, ``Online iterative learning compensation method based on model prediction for trajectory tracking control systems,'' {\em IEEE Trans. Ind. Informatics}, vol.~18, pp.~415--425, Jun. 2022.

\bibitem{AI_TO3}
Z.~Meng, C.~She, G.~Zhao, and D.~De~Martini, ``Sampling, communication, and prediction co-design for synchronizing the real-world device and digital model in metaverse,'' {\em IEEE J. Sel. Areas Commun.}, vol.~41, pp.~288--300, Nov. 2023.

\bibitem{SC}
X.~Luo, H.-H. Chen, and Q.~Guo, ``Semantic communications: Overview, open issues, and future research directions,'' {\em IEEE Wireless Commun.}, vol.~29, pp.~210--219, Jan. 2022.

\bibitem{TO1}
D.~Gündüz, Z.~Qin, I.~E. Aguerri, H.~S. Dhillon, Z.~Yang, A.~Yener, K.~K. Wong, and C.-B. Chae, ``Beyond transmitting bits: Context, semantics, and task-oriented communications,'' {\em IEEE J. Sel. Areas Commun.}, vol.~41, pp.~5--41, Nov. 2023.

\bibitem{SA1}
H.~Xie, Z.~Qin, G.~Y. Li, and B.-H. Juang, ``Deep learning enabled semantic communication systems,'' {\em IEEE Trans. Signal Process.}, vol.~69, pp.~2663--2675, Apr. 2021.

\bibitem{SA2}
Z.~Weng and Z.~Qin, ``Semantic communication systems for speech transmission,'' {\em IEEE J. Sel. Areas Commun.}, vol.~39, pp.~2434--2444, Jun. 2021.

\bibitem{SA4}
M.~Kountouris and N.~Pappas, ``Semantics-empowered communication for networked intelligent systems,'' {\em IEEE Commun. Mag.}, vol.~59, pp.~96--102, Jun. 2021.

\bibitem{TOSA3}
Z.~{Wang}, Y.~{Deng}, and A.~H. {Aghvami}, ``Task-oriented and semantics-aware communication framework for augmented reality,'' {\em arXiv e-prints}, p.~arXiv:2306.15470, June 2023.

\bibitem{AoI2}
R.~D. Yates, Y.~Sun, D.~R. Brown, S.~K. Kaul, E.~Modiano, and S.~Ulukus, ``Age of information: An introduction and survey,'' {\em IEEE J. Sel. Areas Commun.}, vol.~39, pp.~1183--1210, Mar. 2021.

\bibitem{AoI1}
G.~J. Stamatakis, N.~Pappas, A.~Fragkiadakis, and A.~Traganitis, ``Semantics-aware active fault detection in {IoT},'' in {\em Proc. 2022 20th Int. Symp. Model. Optim. Mobile, Ad hoc, Wireless Networks (WiOpt)}, pp.~161--168, Sep. 2022.

\bibitem{VoI1}
W.~Yang, H.~Du, Z.~Q. Liew, W.~Y.~B. Lim, Z.~Xiong, D.~Niyato, X.~Chi, X.~Shen, and C.~Miao, ``Semantic communications for future internet: Fundamentals, applications, and challenges,'' {\em IEEE Commun. Surv. Tutorials}, vol.~25, pp.~213--250, Nov. 2023.

\bibitem{VoI2}
A.~Kosta, N.~Pappas, A.~Ephremides, and V.~Angelakis, ``Age and value of information: Non-linear age case,'' in {\em Proc. 2017 IEEE Int. Symp. Inf. Theory (ISIT)}, pp.~326--330, Jun. 2017.

\bibitem{VoI3}
E.~Fountoulakis, N.~Pappas, and M.~Kountouris, ``Goal-oriented policies for cost of actuation error minimization in wireless autonomous systems,'' {\em IEEE Commun. Lett.}, pp.~1--1, Jun. 2023.

\bibitem{SA3}
D.~Huang, F.~Gao, X.~Tao, Q.~Du, and J.~Lu, ``Toward semantic communications: Deep learning-based image semantic coding,'' {\em IEEE J. Sel. Areas Commun.}, vol.~41, pp.~55--71, Nov. 2023.

\bibitem{TOSA2}
A.~{Gupta}, P.~{Zhang}, G.~{Lalwani}, and M.~{Diab}, ``{CASA-NLU}: Context-aware self-attentive natural language understanding for task-oriented chatbots,'' {\em arXiv e-prints}, p.~arXiv:1909.08705, Sept. 2019.

\bibitem{TOSA4}
H.~Sun, J.~Bao, Y.~Wu, and X.~He, ``{M}ars: Modeling context {\&} state representations with contrastive learning for end-to-end task-oriented dialog,'' in {\em Findings of the Association for Computational Linguistics: ACL 2023} (A.~Rogers, J.~Boyd-Graber, and N.~Okazaki, eds.), (Toronto, Canada), pp.~11139--11160, Association for Computational Linguistics, July 2023.

\bibitem{TOSA5}
Y.~Xu, H.~Zhou, and Y.~Deng, ``Task-oriented semantics-aware communication for wireless {UAV} control and command transmission,'' {\em IEEE Commun. Lett.}, pp.~1--1, Jun. 2023.

\bibitem{Proactive}
Y.~Liu, Y.~Deng, M.~Elkashlan, A.~Nallanathan, and G.~K. Karagiannidis, ``Analyzing grant-free access for {URLLC} service,'' {\em IEEE J. Sel. Areas Commun.}, vol.~39, pp.~741--755, Aug. 2021.

\end{thebibliography}
\begin{IEEEbiography}[{\includegraphics[width=1in, height=1.25in,clip,keepaspectratio]{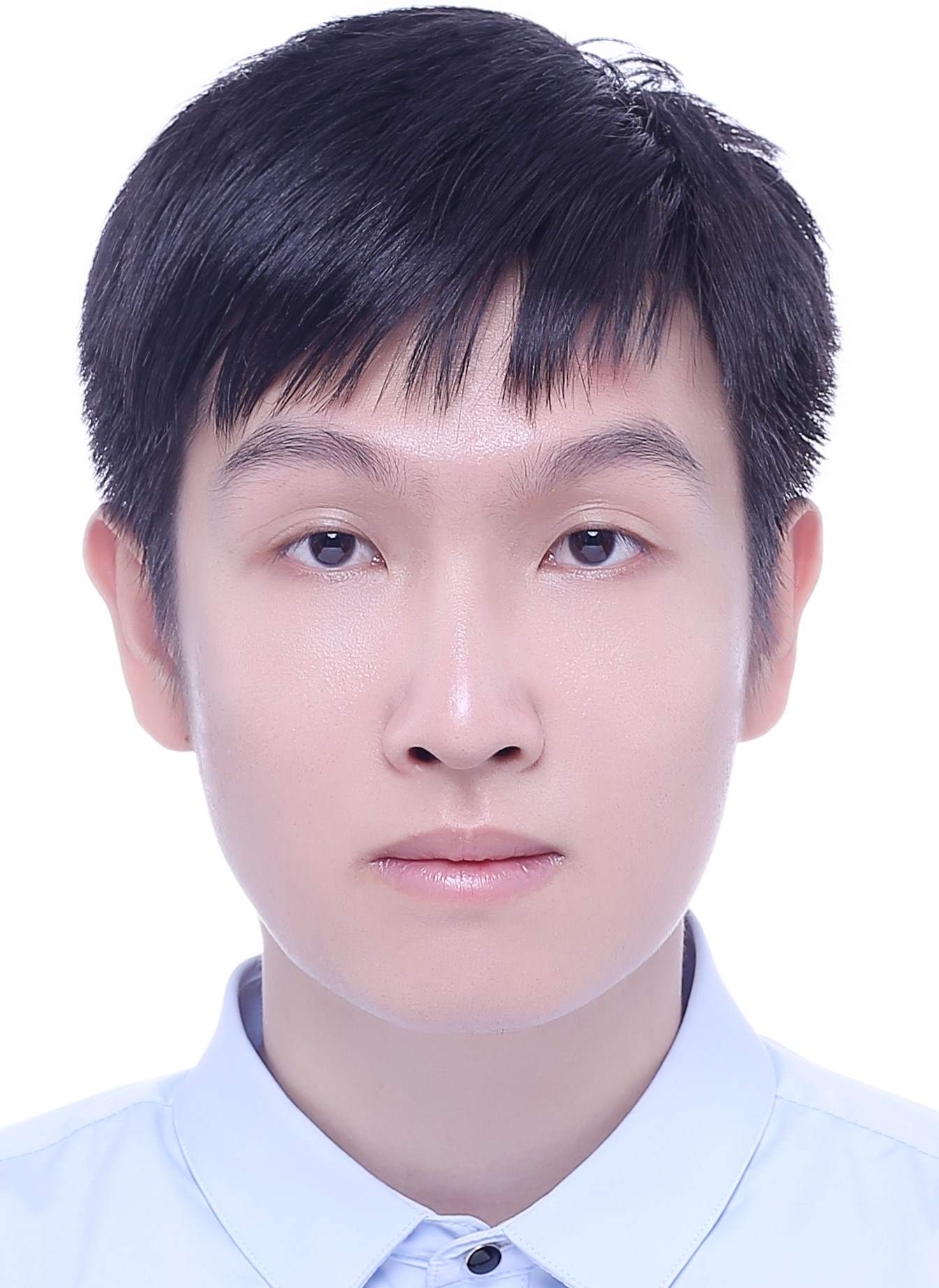}}]{Wenchao Wu} (Student Member, IEEE) is currently pursuing the Ph.D. degree with the Center for Telecommunications Research (CTR), King’s College London. His current research interests include wireless semantic communication for robotics.
\end{IEEEbiography}
\begin{IEEEbiography}[{\includegraphics[width=1in, height=1.25in,clip,keepaspectratio]{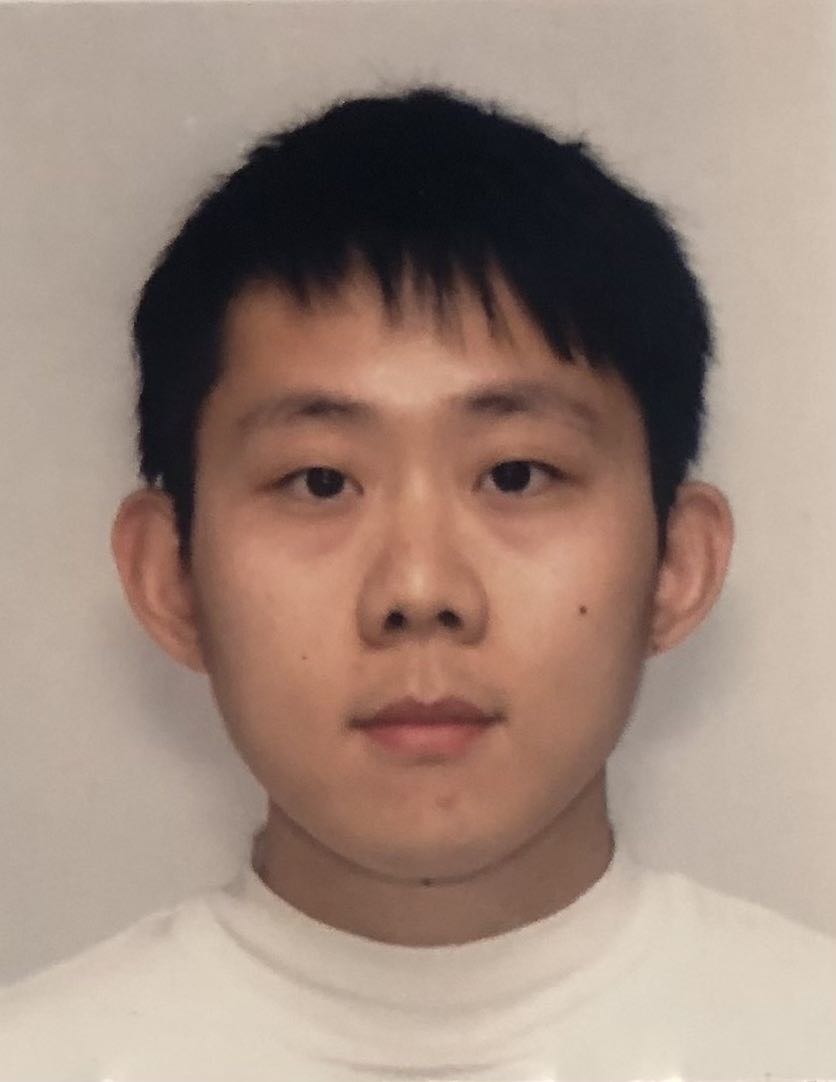}}]{Yuanqing Yang} is a PhD student at the Center for Robotic Research (CORE), King's College London. His current research focuses on applying reinforcement learning to robotic applications and intelligent robot systems.
\end{IEEEbiography}
\begin{IEEEbiography}[{\includegraphics[width=1in, height=1.25in,clip,keepaspectratio]{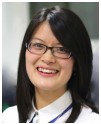}}]{Yansha Deng} (Senior Member, IEEE) received the Ph.D. degree in electrical engineering from the Queen Mary University of London, U.K., in 2015. From 2015 to 2017, she was a Post-Doctoral Research Fellow with King’s College London, U.K., where she is currently a Reader (an Associate Professor) with the Department of Engineering. Her current research interests include molecular communication and machine learning for 5G/6G wireless networks. She was a recipient of the Best Paper Awards from ICC in 2016 and GLOBECOM in 2017 as the first author and IEEE Communications Society Best Young Researcher Award for the Europe, Middle East, and Africa Region in 2021. She also received the exemplary reviewers of the IEEE TRANSACTIONS ON COMMUNICATIONS in 2016 and 2017 and IEEE TRANSACTIONS ON WIRELESS COMMUNICATIONS in 2018. She has also served as a TPC Member for many IEEE conferences, such as IEEE GLOBECOM and ICC. She is currently an Associate Editor of IEEE TRANSACTIONS ON COMMUNICATIONS, IEEE COMMUNICATIONS SURVEYS AND TUTORIALS, IEEE TRANSACTIONS ON MACHINE LEARNING IN COMMUNICATIONS AND NETWORKING, and IEEE TRANSACTIONS ON MOLECULAR, BIOLOGICAL AND MULTI-SCALE COMMUNICATIONS; a Senior Editor of the IEEE COMMUNICATION LETTERS; and the Vertical Area Editor of IEEE Internet of Things Magazine.
\end{IEEEbiography}
\begin{IEEEbiography}[{\includegraphics[width=1in, height=1.25in,clip,keepaspectratio]{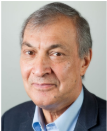}}]{A. Hamid Aghvami} (Life Fellow, IEEE) joined King’s College London, an Academic Staff, in 1984. In 1989, he was promoted to a Reader, and was promoted to a Professor of telecommunications engineering in 1993. He was a Visiting Professor at NTT Radio Communication Systems Laboratories in 1990, a Senior Research Fellow at BT Laboratories from 1998 to 1999, and was an Executive Advisor of Wireless Facilities Inc., USA, from 1996 to 2002. He was the Director of the Centre from 1994 to 2014. He is currently the Founder of the Centre for Telecommunications Research, King’s College London. He is the Chairperson of Advanced Wireless Technology Group Ltd. He is also the Managing Director of Wireless Multimedia Communications Ltd., his own consultancy company. He is also a Visiting Professor at Imperial College London. He carries out consulting work on digital radio communications systems for British and international companies. He has published over 580 technical journals and conference papers, filed 30 patents, and given invited talks and courses the world over on various aspects of mobile radio communications. He leads an active research team working on numerous mobile and personal communications projects for fourth and fifth-generation networks; these projects are supported both by the government and industry. He was a member of the Board of Governors of the IEEE Communications Society from 2001 to 2003, was a Distinguished Lecturer of the IEEE Communications Society from 2004 to 2007, and has been a member, the Chairperson, and the Vice Chairperson of the technical program and organizing committees of a large number of international conferences. He is a fellow of the Royal Academy of Engineering and a fellow of the IET. He was awarded the IEEE Technical Committee on Personal Communications (TCPC) Recognition Award in 2005 for his outstanding technical contributions to the communications field, and for his service to the scientific and engineering communities. In 2009, he was awarded a Fellowship of the Wireless World Research Forum in recognition of his personal contributions to the wireless world, and for his research achievements as the Director of the Centre for Telecommunications Research, King’s College London. He is also the Founder of the International Symposium on Personal Indoor and Mobile Radio Communications (PIMRC), a major yearly conference attracting some 1,000 attendees.
\end{IEEEbiography}

%





\end{document}